\documentclass[11pt,a4paper]{article}
\usepackage[fontset=fandol,heading=false,scheme=plain]{ctex}
\usepackage[margin=2.4cm]{geometry}
\usepackage{amsmath,amssymb}
\usepackage{booktabs,graphicx,array,multirow,adjustbox}
\usepackage{pgfplots,pgfplotstable}
\pgfplotsset{compat=1.18}
\usepackage[numbers,sort&compress]{natbib}
\usepackage{xcolor}
\usepackage[colorlinks=true,linkcolor=blue!60!black,citecolor=blue!60!black,urlcolor=blue!60!black]{hyperref}
\usepackage{caption}
\xeCJKsetup{CJKmath=true}
\newcommand{\paperTitle}{One Byte, One Token: Byteifying a 27B Hybrid-Attention LLM with Self-Generated Data}
\newcommand{\paperTitleZh}{一字节一 token:用模型自产语料把 27B 混合注意力模型改造成纯字节模型}
\newcommand{\paperAuthor}{Liwei Zhang \qquad ZhiMing Tong \qquad Mochen Guan \qquad Linzhe Li$^{*}$\\[0.3em]
\normalsize Landi Lab \qquad $^{*}$Corresponding author\\
\normalsize\texttt{z@metiy.xyz} \quad \texttt{mynameistong@163.com} \quad \texttt{gmc@antiy.cn} \quad \texttt{lilinzhe@antiy.cn}}
\newcommand{\paperAuthorZh}{张栗伟(Liwei Zhang) \quad 童志明(ZhiMing Tong)\\[0.2em]
关墨辰(Mochen Guan) \quad 李林哲(Linzhe Li)$^{*}$\\[0.4em]
\normalsize 澜砥实验室(Landi Lab) \qquad $^{*}$通讯作者\\
\normalsize\texttt{z@metiy.xyz} \quad \texttt{mynameistong@163.com} \quad \texttt{gmc@antiy.cn} \quad \texttt{lilinzhe@antiy.cn}}

\title{\paperTitle}
\author{\paperAuthor}
\date{November 2026}

\begin{document}

\maketitle
\begin{abstract}
Byte-level BPE fragments Chinese text (for Qwen3.8 on Chinese Wikipedia, 3.84 bytes per token on average, with 64\% of tokens being single characters) and hides spelling where the model cannot see it. We ask whether an already-trained 27B subword model can be converted directly into a pure byte-level model—one byte, one token—without adding a patching structure or a local encoder, while keeping most of its ability. Starting from Qwen3.8-27B (48 Gated DeltaNet layers + 16 full-attention layers), we shrink the vocabulary from 248,320 to 384 entries (the 256 byte symbols already present in the BPE vocabulary plus special ids), leave every other weight untouched, and continue training in a two-stage curriculum (encoding first, then reading) on 4.77 GB of supervised bytes: all dialogue/behavioral supervision is generated on-policy by the original model itself (character dictionary, word dictionary, general dialogue, coding sessions), and knowledge reading uses human-written encyclopedia text (54\% of supervised bytes). Training takes 178.6 h on 8$\times$H200 (about 1,430 GPU-hours). Compared on the same GPU with the same serving stack, the byte model's gap to the original falls within the paired bootstrap interval on MATH-500, HumanEval+, MMLU-Pro and non-thinking GSM8K, and it loses 4–6 points on C-Eval, thinking-mode GSM8K and non-thinking IFEval. Without thinking, it leads across the board on character-level tasks (CUTE 75.2$\rightarrow$84.0, EXECUTE-zho 81.6$\rightarrow$93.9, our Chinese character suite 54.4$\rightarrow$64.0), but lookup knowledge about Chinese characters such as pronunciation and stroke count drops, and thinking does not recover it. The cost is throughput: per-step speed is identical, but each step yields 3–5$\times$ less output; on the same GPU, Chinese characters per second is about 1/4 of the original, and prefill at 64K bytes is 6.5$\times$ slower. We also document four failure modes specific to byte models (greedy runaway, sampling breaking UTF-8, penalty terms being ineffective on a small vocabulary, repetition) and their decoding-side and training-side fixes.

\end{abstract}
\noindent\small This document contains the full paper in English followed by the full paper in Chinese (p.~\pageref{sec:zh}). Both versions describe the same experiments and report the same numbers.
\normalsize
\section{Introduction}
A subword tokenizer is a design decision frozen before training. For Chinese, Qwen3.8's byte-level BPE averages 3.84 bytes per token on Chinese Wikipedia, 64\% of tokens are single characters (\texttt{tokenizer-eval/}), and a word is often split into several fragments; for every language, spelling, character positions and the individual digits of numbers are invisible to the model, so character-level operations must rely on "thinking" to break a word apart before acting on it (Section 5.4). Once the vocabulary is fixed, all subsequent training and serving are bound to it.

The most direct alternative is a byte model: 1 byte = 1 token, a vocabulary of 256 plus special symbols. Training a byte model from scratch is too expensive; the two existing practical routes each carry a cost—keep a patching structure (local encoder + patches, as in BLT and Bolmo), in which case the backbone still sees patched vectors; or do cross-tokenizer distillation through a byte-level interface (still judged an open problem at 1B–8B). This paper takes a third route: \textbf{convert an already-trained 27B subword model, as is, into a pure byte-level model without patching}, replacing only the vocabulary and embeddings and leaving the backbone untouched.

Our approach is: shrink the vocabulary to 384 (the BPE vocabulary already contains 256 byte symbols), then continue training with a two-stage curriculum—first learn the "character $\leftrightarrow$ bytes" encoding from the character dictionary, the word dictionary and dialogue, then read human-written encyclopedia text (Section 3). All dialogue/behavioral supervision is generated by the original Qwen3.8 itself (on-policy self-distillation, with per-turn independent re-answering and supervising only the current turn); knowledge reading uses human-written text (54\% of supervised bytes). The whole run uses 4.77 GB of supervised bytes and takes 178.6 h on 8$\times$H200.

This paper sets out to answer only four questions, which are also its four contributions:

\begin{enumerate}
\item \textbf{How much is retained} (Sections 5.1, 5.2): a two-column comparison on the same GPU, the same serving stack and the same decoding protocol. In non-thinking mode, 4 of 7 items show no detectable difference within the paired bootstrap interval, while C-Eval and non-thinking IFEval drop 4–6 points; in thinking mode, GSM8K $-$4.0 and C-Eval $-$4.5.
\item \textbf{What bytes bring} (Section 5.4): without thinking, character-level operations lead across the board (CUTE +8.8, EXECUTE-zho +12.2, our Chinese character suite +9.7), with larger margins in Chinese than in English; but lookup knowledge about Chinese characters such as pronunciation and stroke count drops, and thinking does not recover it.
\item \textbf{What it costs} (Section 5.6): per-step speed is identical, but each step yields 3–5$\times$ less output; on the same GPU, Chinese characters per second falls to 1/4, and prefill at 64K bytes is 6.5$\times$ slower; a draft head is the way to recover speed, but it does not fit on a single 48G GPU.
\item \textbf{Failure modes specific to byte models, and their fixes} (Section 6): greedy runaway, byte-wise sampling breaking UTF-8, penalty terms being ineffective on a small vocabulary, and thinking that cannot find its way out; together with the corresponding decoding guard and post-training methods.
\end{enumerate}
We do not claim that the byte model is stronger. The honest conclusion is: \textbf{at 27B scale and a continued-training budget of about 1e21 FLOPs, byteification without patching can match most of the original model's ability, gaining character-level operations at the cost of throughput and part of the character-level knowledge.} The evaluation scripts, our Chinese character suite, and the scoring / bootstrap code for every evaluation in this paper are available at \texttt{https://huggingface.co/datasets/leiwhere/ld3-byte-eval}; the weights and the training corpus are not released at this time.

\section{Related Work}
\textbf{Byte- and character-level models trained from scratch.} ByT5 \citep{xue2022byt5} feeds T5 raw UTF-8 bytes and is an early representative of "flat bytes"; CANINE \citep{clark2022canine} and Charformer \citep{tay2022charformer} shorten the character sequence with downsampling or differentiable soft subwords; MEGABYTE \citep{yu2023megabyte}, SpaceByte \citep{slagle2024spacebyte}, BLT \citep{pagnoni2025blt} and H-Net \citep{hwang2026hnet} build patching into the architecture—fixed-length, whitespace-based, entropy-based or end-to-end learned boundaries—and run the backbone over the patches. MambaByte \citep{wang2024mambabyte} and EvaByte \citep{zheng2025evabyte} keep flat bytes but respectively switch to an SSM backbone or train a 6.5B model from scratch. All of these works train from scratch; our starting point is different: whether a subword model that has already been trained on tens of trillions of tokens can be converted directly into a byte model without retraining the backbone and without adding a patching structure.

\textbf{Converting subword models into byte models.} BLT reports an experiment that initializes the global Transformer from Llama 3.1 8B (220B tokens), keeping patching and a newly trained local encoder/decoder \citep{pagnoni2025blt}. Bolmo \citep{minixhofer2025bolmo} does this systematically: Olmo 3 7B is converted into a byte-level "latent tokenizer language model", with bytes pooled into patches and an mLSTM local encoder/decoder, about 49B tokens over two stages in total (39.3B in the second stage); patching is likewise kept, and the backbone runs over patches. \citet{bao2026distillbyte} uses progressive distillation to align a token teacher's representations to an H-Net-style hierarchical byte student ($\leq$ 4B). \citet{singh2026bld} uses bytes only as the interface for cross-tokenizer distillation, with the student still a subword model, and reports that consistent gains at 1B–8B "remain hard to obtain". The inference-time method of \citep{hayase2025bytesampling} involves no training and turns any BPE model into a byte-level sampler, while the model internally still computes over subwords. Compared with these, this paper (i) does no patching at all, with the backbone taking one step per byte; (ii) introduces no external teacher, with supervision coming from the original model's own outputs and human-written text; (iii) works at 27B scale with a focus on Chinese, and reports the efficiency cost on the same GPU. The cost of not patching is that the sequence length is not compressed, which Section 5.6 reports as is.

\textbf{Vocabulary transfer.} ZeTT \citep{minixhofer2024zett}, FOCUS \citep{dobler2023focus} and TokAlign \citep{li2025tokalign} study how to initialize new embeddings and recover in a few steps when switching to a different subword tokenizer. This paper is the extreme case: the new vocabulary has only the 256 byte symbols left, and these 256 rows are already in the original embedding table (GPT-2-style byte-level BPE \citep{radford2019gpt2}), so no initialization trick is needed and all of the difficulty shifts to continued training.

\textbf{Self-generated data and on-policy training.} Self-Instruct \citep{wang2023selfinstruct}, STaR \citep{zelikman2022star} and RFT \citep{yuan2023rft} train a model on its own (filtered) outputs; SDFT \citep{yang2024sdft} has the seed model rewrite task answers to mitigate forgetting; \citet{chen2026retaining} attributes the reduced forgetting of RL to on-policy data. Our corpus recipe (Section 3.2) and post-training (Section 3.4) rest on the same observation—dialogue supervision must be the target model's own current output—and apply it to the setting of vocabulary replacement, a drastic distribution shift; DPO \citep{rafailov2023dpo} is used as the means of controlling repetition, with length normalization.

\textbf{Base model and benchmarks.} Qwen3.8-27B \citep{qwen2026qwen38} is a dense model mixing Gated DeltaNet \citep{yang2025gdn} and full attention at a 3:1 ratio (see \citep{qwen2025qwen3next,qiu2026qwen38next} for descriptions of the same architecture family), and natively carries one MTP layer \citep{gloeckle2024mtp,deepseek2024v3}. Character-level tasks use CUTE \citep{edman2024cute} and its multilingual extension EXECUTE \citep{edman2025execute}.

\section{Method}
\subsection{Vocabulary Shrinking}
The tokenizer of Qwen3.8-27B is a GPT-2-style byte-level BPE: the vocabulary already contains 256 single-byte symbols, and every multi-byte token is formed by merging them. Converting to a pure byte model therefore requires no new embeddings at all; we only delete every merge entry and keep the 256 byte rows plus the special symbols that the chat template depends on. The shrunk vocabulary has 384 entries: bytes 0–255 ordered by byte value, followed by 35 named special symbols such as \texttt{\textless{}\textbar{}endoftext\textbar{}\textgreater{}}=256, \texttt{\textless{}\textbar{}im\_start\textbar{}\textgreater{}}=257 and \texttt{\textless{}think\textgreater{}}=280, with the remaining 93 positions as placeholders (full table in Appendix B). The tokenizer contains no merge rules whatsoever: one UTF-8 byte is one token.

The backbone is left untouched: 64 layers = 48 GatedDeltaNet layers + 16 full-attention layers (one every 4 layers), hidden size 5120, \texttt{model\_type} still \texttt{qwen3\_5}, so off-the-shelf inference frameworks load it directly. Parameters change only at the two ends: the embedding and the lm\_head each shrink from 248,320 $\times$ 5,120 $\approx$ 1.27B to 384 $\times$ 5,120 $\approx$ 2M, and the language-model part goes from 26.9B to 24.4B; the 0.46B vision tower is frozen as is. The shrinking script selects rows by token name rather than by id offset; the output is checked item by item against the special-id table and the md5 of the chat template (two easily missed errors: the vision token ids in the config, and the reordering of special symbols by newer versions of transformers).

The cost is longer sequences: the original tokenizer averages 3.84 bytes/token on Chinese and 4.35 on English, so byte-level sequences are about 4$\times$ as long, and a 16K-byte window corresponds roughly to 4k BPE tokens. In Chinese text, only 15.9\% (digits, punctuation, newlines) already falls on single-byte tokens, and 99.4\% of Chinese characters never pass through the byte path in the original model, so the zero-adaptation model can read but reads poorly: the untrained shrunk checkpoint has four-domain bpb of zh 3.02 / en 4.43 / code 2.54. These numbers serve only to check that the shrinking was done correctly (they are of the same order as the zero-adaptation numbers of the first 27B pilot) and are not taken as a measure of starting ability.

\subsection{Corpus: Self-Generated Dialogue Supervision, Human-Written Text for Knowledge Reading}
Across its two-stage curriculum, LD3-base read 7.08 GB in total, of which 4.77 GB were supervised bytes. Supervised bytes by source: wiki text 2.56 GB (54\%), word dictionary 1.61 GB (34\%), character dictionary 0.30 GB (6\%), general dialogue 0.16 GB, coding sessions 0.14 GB. To be clear from the outset: \textbf{the wiki, which accounts for 54\% of supervised bytes, is human-written Wikipedia text}, read in pretraining fashion, concatenated end to end and cut into windows; it is not model-generated. What is self-generated is the other four categories of dialogue / behavioral supervision. The human text is responsible for knowledge reading; the self-generated dialogue is responsible for "the model speaking in its own words in bytes".

All dialogue supervision is generated by the original Qwen3.8-27B itself (on-policy), on the basis of two sets of controlled comparisons. Among the six "less thinking" training comparisons on the Ornith pilot lines, only training the model on its own current outputs came at zero cost (3.2 passes over its own correct samples, accuracy unchanged at 49\%), whereas all four recipes that used data written by others lost 5–14 points, regardless of content quality; "current" is strict enough that outputs from one training stage earlier no longer count. A three-arm comparison in cross-vocabulary distillation points to the same thing: supervising prompt prefixes written by others made code bpb 26\% worse and en 16\% worse, and the damage disappeared once supervision was restricted to the current turn's generation. Hence, in this paper, CE is computed only on assistant turns (including thinking); multi-turn sessions use \textbf{per-turn independent re-answering}, with the prefix kept verbatim and supervising only the current turn.

The four self-generated corpora are: (1) the \textbf{character dictionary}, 12,126 Chinese characters (8,105 standard characters + 4,021 traditional characters whose written form differs from the simplified one), with readings, senses and simplified–traditional relations taken from Unihan / CC-CEDICT / OpenCC, three dialogue turns per character (entry $\rightarrow$ translate the whole entry into the other language $\rightarrow$ follow-up question); it is the most direct explicit supervision of "one character $\leftrightarrow$ three bytes", and so comes first. (2) The \textbf{word dictionary}, 141,361 entry seeds $\times$ 4 turns. The writing rule for both dictionaries is "facts come from the dictionary, the wording is written by the teacher": prompts are natural dialogue, format requirements are placed only in the unsupervised user turn, and the persona is spread over 33 variants so that the writing style is not pulled toward a single persona. (3) \textbf{General dialogue}, 60,549 sessions after deduplicating six public session pools, 1.13 million assistant turns in total, 91\% of which are agent turns with tool calls. (4) \textbf{Human coding sessions}, two public datasets (6,772 + 1,064 sessions) re-answered twice per turn, 88,743 turns in total. All generation was done with thinking on and the official sampling parameters (temperature 1.0 / top\_p 0.95 / top\_k 20).

The amount generated far exceeds the amount used in training: more than 300,000 turns of general dialogue were generated locally, yet only 0.16 GB of supervised bytes entered the curriculum, because prefixes are long and only the last turn is supervised. Local generation totaled about 94.9 h $\approx$ 760 H200 GPU-hours; for the part that went through hosted APIs (2,048 essays, about 11,800 logic problems, coding sessions) the compute is not visible and only call volumes are reported. Essays and logic problems \textbf{did not enter the two-stage curriculum of LD3-base}: the logic problems (reasoning-gym, 99 categories, 10,560 problems, 9,618 answered correctly) are used in Section 3.4, and the essays (two batches, about 2,560 in total) were only generated and never entered any training pack.

\subsection{Two-Stage Curriculum: Learn the Encoding First, Then Read}
All training is full-parameter with ZeRO-3, and checkpoints store weights only (47 GB). Both stages use a constant learning rate; between stages only the weights are carried over (optimizer reset, 5 warmup steps), and no bpb regression was observed in practice. Decisions were made by looking jointly at four-domain byte bpb, a 30-problem greedy probe and a held-out dialogue set; no decision was made on any one of them alone.

\textbf{M1 (encoding stage)} (64K-byte window, 2,077 steps, lr 3e-6, 47.5 h): character dictionary : word dictionary : general dialogue mixed 1:1:1 by supervised bytes, with the character dictionary used in full for one pass, the word dictionary sampled for character coverage, and general dialogue restricted to sessions that fit entirely within 64K. Actually read: 1.143 GB; supervised: 0.616 GB (the "0.28 GB each, 0.85 GB total" in the training records is the sampling target; this paper uses the pack statistics). The bpb trajectory shows that "Chinese is the new skill": en reaches 0.58 within 100 steps and stays there, while zh goes from 3.02 through 1.12 at step 400 to 0.900 at the end. But at the end of M1 the model can read without being able to use what it reads: the probe scores 29/30, yet "English $\rightarrow$ Chinese" terminology scores 0/12 while the English paths score 11–12/12—the knowledge is there; what is broken is the link between Chinese word forms and concepts.

\textbf{M2 (reading stage)} (16K-byte window, 6,761 steps, lr 5e-6, three continuation segments totaling 131 h): wiki text : general corpus : word dictionary = 1:1:1 mixed \textbf{by item count}, which in supervised bytes is wiki 54\% / word dictionary 34\% / dialogue 6\% (Section 3.2), because dialogue prefixes carry no loss. Actually read: 5.932 GB; supervised: 4.156 GB. After M2, "English $\rightarrow$ Chinese" terminology went from 0/12 to 12/12 (already reached at 3,000 steps), the probe to 30/30, and GSM8K in dialogue form to 294 of 300 problems; MMBench-CN rose from 65.2\% to 72.2\%, so text-only training did not harm the frozen vision tower.

Both stages \textbf{mix, never sequence}: on the pilot lines, the recipe "train on the narrow corpus first, then patch with normal data afterwards" pushed the behavior back, but the lost abilities did not return; so M1 mixes in general dialogue from the very first step, and in M2 the wiki is likewise mixed together with dialogue.

\subsection{Byte-Specific Post-Training: Methods and Negative Lessons}
LD3-base received no post-training of any kind. This section records the methods and conclusions of a post-training chain built on top of it (denoted M3); the weights of this chain have been deleted and only the logs remain, and it does not enter the main table.

The trigger was an observation after deployment: 46\% of LD3-base's self-generated answers contained 8-gram repetition (another commercial model on the same questions: 6\%), and it did not follow the writing style guide written into the system prompt. Since prompting was ineffective, the only remaining option was to train it in. M3's goals, in order of priority, were avoiding repetition, concise and correct thinking, and working according to the style guide; the chain has four steps, all on-policy:

\begin{enumerate}
\item \textbf{Style-guide SFT}: 10.5K style-guide general dialogues + about 5K style-guide logic problems (all correctly answered ones kept) + 4K replay, 16K window, lr 5e-6 with WSD, 259 steps.
\item \textbf{Self-sampling with the decoding guard off}: prompts are taken from the training pack as the prefix "up to just before the last-turn answer", 4 samples per prompt, official sampling parameters, 8,192 bytes per sample cap. The decoding guard is turned off during sampling so that repetition is exposed; a sample cut off by the guard, once it enters the training data, teaches the model to write \texttt{\textless{}/think\textgreater{}} in the answer segment.
\item \textbf{Mining samples and RFT}: clean = stopped on its own, contains \texttt{\textless{}/think\textgreater{}}, 8-gram repetition rate $\leq$ 0.03, no whole-sentence repetition or stutter, style-lint score $\geq$ 80; bad = hitting the output cap, or repetition rate $\geq$ 0.10, or whole-sentence repetitions $\geq$ 2; for logic problems the chosen is "correct with the shortest thinking". Clean samples are first used for one round of RFT (lr 3e-6).
\item \textbf{Length-normalized DPO}: at most one pair per prompt, chosen = the clean sample with the highest lint score, rejected = the bad sample with the highest repetition rate. Reward r = $\beta$$\cdot$(log p $-$ log p\_ref)/len, loss L = $-$log $\sigma$(r\_c $-$ r\_r $-$ $\gamma$) + $\alpha$$\cdot$($-$log p\_c/len\_c), $\beta$ 2.0 / $\gamma$ 0.5 / $\alpha$ 0.2, lr 5e-7. The normalization is necessary: in 99\% of pairs the rejected is longer than the chosen (median lengths 634 vs 4,918 bytes).
\end{enumerate}
Three rounds yielded three negative lessons:

\begin{itemize}
\item \textbf{Skipping RFT and going straight to DPO hurts precise reasoning.} 61,600 self-sampled responses, bad-sample rate 11.9\%; direct DPO (4,460 pairs, 140 steps) brought the bad-sample rate down to 7.7\%, but greedy dialogue GSM8K on 300 problems dropped from 97.7\% to 80.7\%, and mean output length went from 1,198 to 3,681 tokens. (97.7\% is under the protocol of the 300-problem greedy probe with thinking, not the protocol of the main table in Section 5.)
\item \textbf{RFT before DPO brings it back.} Adding a "shortest and correct" RFT before DPO (7,004 pairs, 219 steps) brought GSM8K on 300 problems back to 92.7\%.
\item \textbf{Pushing repetition down further squeezes thinking out.} In the third-round acceptance test the bad-sample rate was only 0.2\%, but of 252 answers with thinking, 211 had thinking shorter than 60 characters, with a median of 28 characters, versus medians of 136 and 210 in the first two rounds. Most of the 0.2\% comes from "not thinking". The root cause is the pairing rule: taking the "shortest" as chosen systematically selects samples without thinking; the rule was later changed so that the chosen's thinking length is $\geq$ 1/4 of the rejected's, the thinking language is the same, and total lengths are as close as possible, so that a pair has no systematic difference other than the defect itself. Do not train when pairs are too few, either: a DPO on 218 pairs for 28 steps had pref\_acc 0.375 and repetition went up instead.
\end{itemize}
\subsection{Deep-Draft MTP}
A byte model emits one byte per step, so its number of decoding steps is about 4$\times$ that of the original model; a multi-token prediction (MTP) draft head is the main means of recovering speed. The Qwen3.5 family natively carries one MTP layer: \texttt{fc[5120, 10240]} concatenates the next-token embedding with the backbone hidden state, followed by 1 full-attention decoder layer, sharing the embedding and lm\_head with the backbone (so the 384-entry vocabulary applies directly), 424.7M parameters; once the backbone is trained, this layer has to be rebuilt separately.

Draft depth is set by the number of Chinese characters, not the 2–3 steps customary for BPE models: a Chinese character is 3 bytes and the continuation bytes within a character are nearly deterministic, so the default is K=9 (3 characters), later raised to K=15 (5 characters); training cost is $\propto$ K, but the MTP layer is only 1/64 of the backbone. The head of LD3-base was trained with the backbone frozen and only the head updated, K=15, 1,500 steps, lr 1e-4; after measurement, a second round is triggered by "fewer than 8 bytes accepted per round or speedup below 1.5$\times$", continuing on-policy on about 100 million bytes (60,000 prompts) written by the model itself, lr 5e-5, keeping whichever head has the higher acceptance length; the two rounds together took about 35 GPU-hours.

Two pitfalls. First, the training-time metric and the serving-time metric differ widely: the K=15 head on the pilot lines had a step-1 hit rate of 0.935, yet in serving it accepted only 5.33 bytes per round and was only 20\% faster, because per-position accuracy decays from 0.93 to 0.40 at position 16. The head of LD3-base, running at K=8 on a single GPU, accepts 5.7 bytes per round, giving 1.8$\times$ at 16 concurrent streams and 3.3$\times$ single-stream (the configuration that could not be re-measured on the single 48G GPU in Section 5.6). Second, once the backbone changes, the old head is void and must be retrained; multi-layer draft models and fine-tuning with a single-step objective both fell short of the 1-layer MTP in our setting.

\subsection{Decoding Guard}
A byte model needs a layer of decoding guard that a BPE model does not; the reasons are developed in Section 6. The guard is implemented as a custom logit processor in the inference framework, with four components:

\begin{itemize}
\item \textbf{Thinking budget and answer budget}: after \texttt{\textless{}think\textgreater{}}, force \texttt{\textless{}/think\textgreater{}} when the budget is reached; after \texttt{\textless{}/think\textgreater{}}, force \texttt{\textless{}\textbar{}im\_end\textbar{}\textgreater{}} at the cap—even after the thought is closed, the answer segment can still loop up to the cap, so both segments need a budget. Serving default is 16,384 bytes for each.
\item \textbf{UTF-8 validity mask}: after a lead byte, only continuation bytes (128–191) are allowed; at a character boundary, bare continuation bytes and invalid lead bytes are forbidden.
\item \textbf{Loop circuit breaker}: force a close / end only when repetition is still ongoing after 1,024 consecutive steps; otherwise \textbf{the logits are not modified}. An early version forbade repetition byte by byte; on the pilot lines the gate was measured to be active on 41\% of decoding steps (repeats within 16 bytes are common anyway in English, LaTeX and code), gaining in thinking mode and losing in non-thinking mode (GPQA thinking 48.0 $\rightarrow$ 60.6, non-thinking 60.6 $\rightarrow$ 53.0).
\end{itemize}
Two rules of use. First, \textbf{the inference framework's overlap scheduling must be turned off}: under overlap scheduling, the output the guard reads lags by one token, so when the model has just emitted a lead byte the guard still believes it is at a character boundary and forbids all legal continuation bytes—the guard itself manufactures the invalid UTF-8 it is meant to prevent; the three-configuration comparison is given in Section 6.2. Second, there is a ceiling on what the guard can recover: on the pilot lines, three guard versions reduced "no extractable answer" from 77 to 41–53 problems and raised the close rate from 126 to 198/198, yet the number correct was 120/198 in every case (the original BPE model: 170); once the decoding side has been squeezed dry, what remains is a capability gap. The guard is therefore a net gain on tasks with "thinking mode + a definite answer", while on tasks that demand long outputs with a specific structure it cuts off structures that were not finished; the guard is turned off when self-sampling training data (Section 3.4). The guard used in the evaluation of Section 4 only blocks invalid byte sequences and over-long outputs and does not change the distribution of legal outputs.

\section{Experimental Setup}
\subsection{Models and Checkpoints}
Only two models are compared:

\begin{itemize}
\item \textbf{orig}: the officially released Qwen3.8-27B (FP8 weights, block-128 recipe), with a byte-level BPE vocabulary of 248,077 entries (embedding table of 248,320 rows).
\item \textbf{LD3-base}: the checkpoint obtained on the same backbone by shrinking the vocabulary to 384 (256 byte symbols + 128 special / reserved ids), reducing the embedding and output head from 1.27B parameters each to about 2M (language-model part 26.9B $\rightarrow$ 24.4B), and continuing training through the two-stage curriculum (Section 3.3), plus an MTP draft head of depth 8 (Section 3.5). Every table in this paper uses this single checkpoint; to align the protocol, it is also quantized to block-128 FP8 with the same recipe.
\end{itemize}
LD3-base has undergone no benchmark-oriented post-training; it is the most direct answer to the question "does byteification preserve the original model's ability". Post-training problems specific to byte models (repetition, greedy runaway) are discussed separately in Section 6 and do not enter the main table.

\subsection{Training Compute}
8$\times$H200, ZeRO-3 full-parameter continued training: M1 (encoding stage) 47.5 h, M2 (reading stage, three segments combined) 131 h, 178.6 h in total for the backbone $\approx$ 1,430 GPU-hours; the MTP head about 35 GPU-hours. 7.08 GB of bytes actually read, 4.77 GB of supervised bytes (loss computed on assistant turns only). Corpus generation is counted separately: self-generated general dialogue / character dictionary / word dictionary / logic problems took about 760 GPU-hours on the local eight-GPU machine (from the generation queue logs), and about 2,190 GPU-hours once external API calls are converted and included (\texttt{E3\_E4.md}). A rough 6ND estimate puts the backbone continued training at about 1e21 FLOPs, the same order of magnitude as Bolmo; \textbf{an order of magnitude less data, but not less compute}, so the main text does not write "N$\times$ cheaper".

\subsection{Evaluation Protocol (E1 / E2)}
Both columns run every benchmark \textbf{on the same RTX 6000 Ada 48G, with the same serving stack} (sglang v0.5.17, FP8, single GPU, no draft, non-streaming, \texttt{--disable-radix-cache}), so that serving-side differences are not counted as model differences. Specifically:

\begin{itemize}
\item \textbf{Two decoding modes}: thinking mode temperature 0.6 / top\_p 0.95, with the content before \texttt{\textless{}/think\textgreater{}} filtered out before scoring; non-thinking mode temperature 0. Both modes are controlled through the chat template's \texttt{enable\_thinking} switch; the template is identical to the official one.
\item \textbf{Output caps aligned as bytes ÷ 4}: the byte model's cap is counted in bytes, and the BPE model gets 1/4 as many tokens (thinking mode 32,768 B / 8,192 tok; non-thinking mode 16,384 B / 4,096 tok; LongBench-v2 5,000 B / 1,250 tok). The measured median output-length ratio bytes / tokens is about 3 (footnote in Section 5.1), so the 4$\times$ protocol is not tighter for the byte model.
\item \textbf{Per-item counts accompany every score}: the number of cap hits (\texttt{stop\_reason = max\_tokens}) and, in thinking mode, the number of items missing \texttt{\textless{}/think\textgreater{}} are reported together, so that "did not finish" is not conflated with "cannot do it".
\item \textbf{Decoding guard}: the byte model's server carries a UTF-8 validity mask, byte budgets for thinking / answer, and a loop breaker (Section 3.6); the BPE model neither needs nor has one. The guard only blocks invalid byte sequences and over-long outputs; it does not change the distribution of valid outputs.
\item \textbf{Datasets}: the full GSM8K test set (1,319), MATH-500 (500), HumanEval+ (164), IFEval (541 prompts / 834 instructions), C-Eval val (1,346), MMLU-Pro (72 questions sampled from each of 14 subjects, 1,008 in total), LongBench-v2 (the 117 questions that fit in the context). Scoring uses evalscope's default scorers, the same code for both columns.
\item \textbf{bpb}: a held-out window set of 4 domains $\times$ 96 windows $\times$ 4,096 bytes; the byte model sums per-byte logprobs directly; the BPE model sums token logprobs and divides by the byte count of the same text.
\item \textbf{Single seed, single run}. For each item we give the per-item bootstrap 95\% interval and the paired $\Delta$ interval on the same questions (2,000 resamples, Appendix Table A2).
\end{itemize}
\subsection{Character-Level Task Protocol (E5)}
\begin{itemize}
\item \textbf{Test sets}: CUTE (leukas/cute, 14 tasks, 400 questions sampled per task), the eng and zho languages of EXECUTE (benjamin/execute; 200 questions per task), and our Chinese character suite of 15 tasks (200 questions per task; traditional/simplified conversion 107 questions each, limited by the number of natural sentences that convert one-to-one). The character list for our suite is taken from the 12,126-character common-use list; traditional/simplified facts from OpenCC \texttt{STCharacters/TSCharacters} (only one-to-one characters kept); pinyin / stroke count / radical from Unihan \texttt{kMandarin/kTotalStrokes/kRSUnicode}; natural sentences from C-Eval question stems (8–40 characters, no ASCII, $\geq$ 80\% Chinese characters, 675 sentences in total). 13,214 questions in total after excluding extra\_*.
\item \textbf{Scoring}: exact match. The "答案:" ("Answer:") prefix, leading/trailing quotes and the final period are stripped first, then whether whitespace is ignored is decided per task type; only the first non-empty paragraph of the response is taken. In thinking mode only the content after \texttt{\textless{}/think\textgreater{}} is considered.
\item \textbf{Decoding}: non-thinking mode at temperature 0 is primary, and both columns run all test sets; thinking mode is run additionally only on all 14 CUTE tasks and our Chinese suite, 100 questions per task (temperature 0.6 / top\_p 0.95). Output cap 1,024 bytes / 256 tokens (thinking mode 16,384 bytes / 4,096 tokens).
\item \textbf{Exclusions}: the three \texttt{extra\_*} splits of EXECUTE-zho (\texttt{extra\_c2g}, \texttt{extra\_contains}, \texttt{extra\_g2c}, component description $\leftrightarrow$ whole character) contain in the public data only inputs of the form "Task: ⿰亻吏" with no task instruction (they were intended for few-shot use); both columns score 0 zero-shot, and they \textbf{are not included in the tables}. EXECUTE-zho therefore has 12 tasks, like eng.
\end{itemize}
\subsection{Efficiency Protocol (E6)}
Same GPU, same serving stack, four configurations: orig, orig + NEXTN draft, LD3-base, LD3-base + NEXTN draft (K = 8). Three quantities: (1) single-stream decoding, 10 Chinese and 10 English essay prompts of about 300 words, reported as \textbf{Chinese characters per second} and \textbf{English characters per second} rather than tokens per second; (2) total throughput at concurrency 16, likewise counted in characters; (3) prefill: Chinese windows of 16,384 and 65,536 bytes, max\_tokens = 1, reporting time to first token (TTFT), median of 3 runs each. The draft configurations need the memory fraction lowered to 0.80 and the context lowered to 65,536 to start on a single GPU; a configuration that fails to start is recorded as such ("does not start on a single GPU").

\section{Results}
\subsection{Main Table: How Much Byteification Preserved}
Table 1 is the main result of this paper. Taking the two modes together:

\begin{itemize}
\item \textbf{Non-thinking mode} (everyday usage): of the 7 items, GSM8K ($-$0.4), MATH-500 (+0.8), HumanEval+ ($-$2.5) and MMLU-Pro ($-$1.3) are within $\pm$2.5; C-Eval $-$4.7; the four IFEval metrics at $-$4.1 to $-$6.2 are the largest loss.
\item \textbf{Thinking mode}: MATH-500 (+1.4) and the four IFEval metrics (+0.2 to +0.6) are level or slightly higher, HumanEval+ $-$1.8, MMLU-Pro $-$1.3; GSM8K $-$4.0 and C-Eval $-$4.5 are the two clear losses.
\item \textbf{The most consistent loss is on the Chinese exam} (C-Eval about $-$4.5 in both modes) \textbf{and on instruction following in non-thinking mode}; \textbf{the most consistent preservation is on mathematics and MMLU-Pro}. LD3-base has had no instruction-type post-training, and part of the non-thinking IFEval gap should be attributed to this (the same item is level in thinking mode), but this paper has no experiment that separates this factor.
\end{itemize}
\begin{table}[htbp]
\centering
\footnotesize
\caption{Main table: scores of orig and LD3-base on the same RTX 6000 Ada 48G with the same serving stack (single seed; cap hits and missing-\texttt{\textless{}/think\textgreater{}} counts in the table notes; paired bootstrap intervals in Appendix Table A2)}
\label{tab:main}
\begin{adjustbox}{max width=\textwidth}
\setlength{\tabcolsep}{3.5pt}
\begin{tabular}{llrrrrrr}
\toprule
Dataset (n) & Metric & orig thinking & \textbf{LD3-base thinking} & $\Delta$ & orig non-thinking & \textbf{LD3-base non-thinking} & $\Delta$ \\
\midrule
GSM8K (1319) & acc & 97.0 & 93.0 & $-$4.0 & 96.4 & 96.0 & $-$0.4 \\
MATH-500 (500) & acc & 94.0 & 95.4 & +1.4 & 92.8 & 93.6 & +0.8 \\
HumanEval+ (164) & pass@1 & 90.2 & 88.4 & $-$1.8 & 92.7 & 90.2 & $-$2.5 \\
IFEval (541) & prompt-strict & 87.8 & 88.2 & +0.4 & 81.9 & 76.2 & $-$5.7 \\
IFEval (541) & prompt-loose & 89.6 & 90.2 & +0.6 & 86.0 & 79.8 & $-$6.2 \\
IFEval (834) & inst-strict & 91.6 & 91.8 & +0.2 & 87.4 & 83.3 & $-$4.1 \\
IFEval (834) & inst-loose & 92.9 & 93.4 & +0.5 & 90.8 & 86.2 & $-$4.6 \\
C-Eval val (1346) & acc & 89.9 & 85.4 & $-$4.5 & 88.1 & 83.4 & $-$4.7 \\
MMLU-Pro (1008) & acc & 80.1 & 78.8 & $-$1.3 & 79.4 & 78.1 & $-$1.3 \\
LongBench-v2 (117)\textdagger{} & acc & 27.4 & 39.3 & +11.9 & 42.7 & 53.8 & +11.1 \\
\bottomrule
\end{tabular}
\end{adjustbox}
\end{table}
Footnotes (to go into the table notes in the main text):

\begin{itemize}
\item \textdagger{} LongBench-v2 is dominated by the output cap in both columns and both modes (cap hits: thinking mode orig 86/117, LD3-base 69/117; non-thinking mode 48/117, 34/117), and 1,250 tokens is tighter for the BPE model than 5,000 bytes is for the byte model (in Chinese 1 token $\approx$ 1.5–2 bytes); \textbf{this row is not taken as evidence of an advantage of byteification}; it only shows that under the same cap the byte model wraps up earlier.
\item Cap hits / missing \texttt{\textless{}/think\textgreater{}} (thinking mode, orig / LD3-base): GSM8K 9/4, MATH-500 28/15, HumanEval+ 13/6, IFEval 28/27, C-Eval 48/31, MMLU-Pro 138/93; non-thinking mode cap hits: GSM8K 7/1, MATH-500 27/9, HumanEval+ 0/0, IFEval 5/22, C-Eval 30/37, MMLU-Pro 61/29. In thinking mode the byte model's cap-hit count is no higher than the original's on every item, so there is no "thinking that never stops"; the original hits the cap on 13.7\% of MMLU-Pro in thinking mode, so its 80.1 may be slightly depressed by the 8,192-token cap.
\item Median output length (thinking mode), orig tokens / LD3-base bytes: GSM8K 289 / 870, MATH-500 607 / 1,382, MMLU-Pro 712 / 2,263, a ratio of about 3.
\item Single seed; per-item bootstrap 95\% intervals (2,000 resamples, paired $\Delta$): the only paired intervals that exclude 0 are GSM8K thinking mode [$-$5.3, $-$2.9], C-Eval thinking [$-$6.3, $-$2.8] / non-thinking [$-$6.5, $-$3.0], and the cap-dominated LongBench-v2; the $\Delta$ intervals of MATH-500, HumanEval+, MMLU-Pro and non-thinking GSM8K all include 0 (full table in Appendix Table A2). Resampling covers only question-sampling variance, not decoding randomness.
\item Wall time (on the same GPU): orig 7 h 44 min for both modes, LD3-base 35 h 57 min for both modes ($\approx$ 4.6$\times$), see Section 5.6.
\end{itemize}
\subsection{Language Modeling: bpb}
Table 2. All four domains are lower by about 2 standard errors, but the window set is the held-out set from LD3 training, drawn from the same distribution as its training data, which the original never saw, so \textbf{this cannot be read as stronger modeling ability}; it only says that after byteification the model returns to, and slightly below, the original's level on its own distribution; the zh\_ood row is the most credible. Figure 1 is the curve over the whole of training: the zero-adaptation starting point (after vocabulary shrinking, before training) zh 3.02 / zh\_ood 3.29 / en 4.43 / code 2.54; at M1 step 100 en has already reached 0.58 while zh is still at 2.02; at the end of M1 zh 0.900 / zh\_ood 1.213 / en 0.575 / code 0.203; during M2 all four domains decline slowly and monotonically, ending at 0.740 / 0.995 / 0.547 / 0.191. The training log is computed on bf16 weights with CE calculated inside the trainer process, whereas E2 uses FP8 weights with logprobs taken through the server; the two readings of the same checkpoint differ by $\leq$ 0.002, so Table 2 and Figure 1 can be read against each other. English returns to the original's level within the first 100 steps of M1, whereas Chinese needed all 7 GB of bytes to catch up—this is what Section 3.3's "Chinese is the new skill" looks like in the loss.

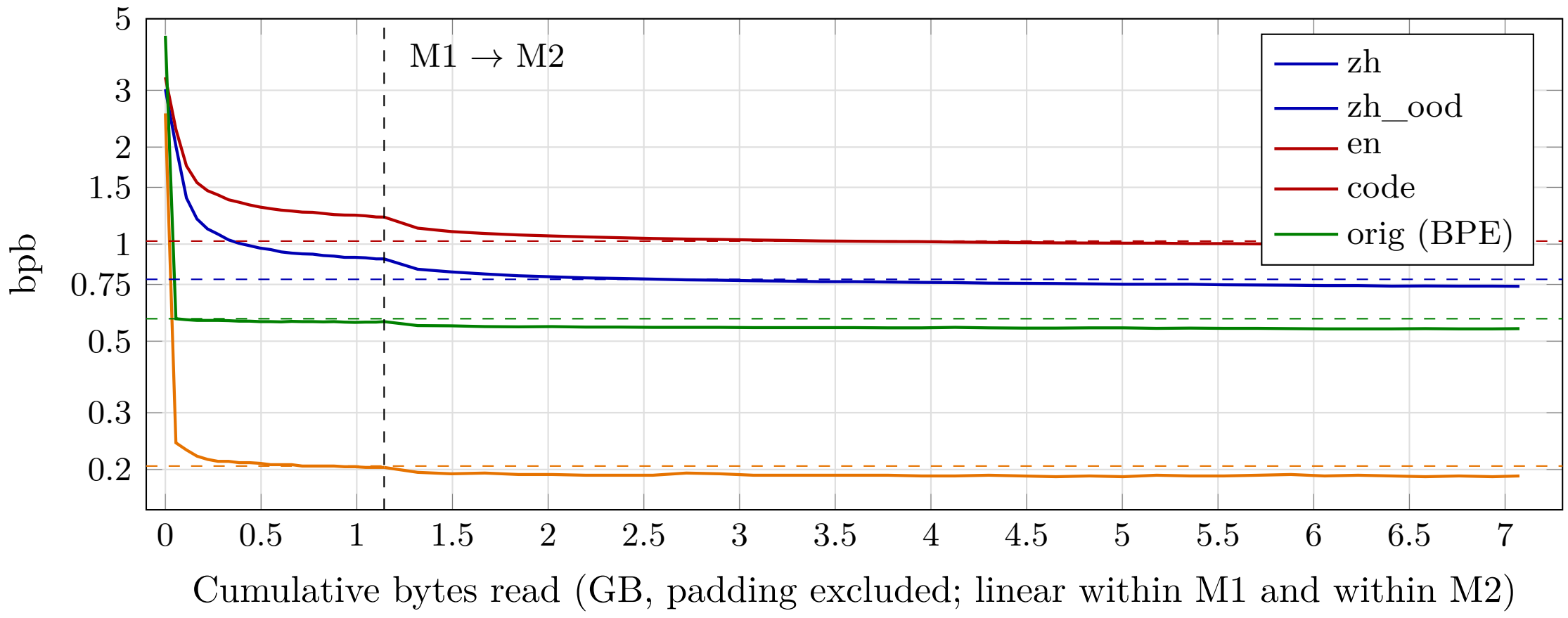
\begin{figure}[htbp]
\centering
\begin{tikzpicture}
\begin{axis}[
  width=0.92\textwidth, height=6.2cm,
  xlabel={Cumulative bytes read (GB, padding excluded; linear within M1 and within M2)},
  ylabel={bpb},
  ymode=log, ymin=0.15, ymax=5,
  log ticks with fixed point,
  ytick={0.2,0.3,0.5,0.75,1,1.5,2,3,5},
  xmin=-0.1, xmax=7.3,
  grid=major, grid style={gray!25},
  legend style={at={(0.98,0.97)},anchor=north east,font=\footnotesize,cells={anchor=west}},
  tick label style={font=\footnotesize}, label style={font=\small},
  every axis plot/.append style={thick},
]
\addplot[blue!70!black]  table[x=bytes,y=zh]     {figs/bpb_m1.dat}; \addlegendentry{zh}
\addplot[blue!70!black]  table[x=bytes,y=zh]     {figs/bpb_m2.dat};
\addplot[red!70!black]   table[x=bytes,y=zh_ood] {figs/bpb_m1.dat}; \addlegendentry{zh\_ood}
\addplot[red!70!black]   table[x=bytes,y=zh_ood] {figs/bpb_m2.dat};
\addplot[green!50!black] table[x=bytes,y=en]     {figs/bpb_m1.dat}; \addlegendentry{en}
\addplot[green!50!black] table[x=bytes,y=en]     {figs/bpb_m2.dat};
\addplot[orange!90!black] table[x=bytes,y=code]  {figs/bpb_m1.dat}; \addlegendentry{code}
\addplot[orange!90!black] table[x=bytes,y=code]  {figs/bpb_m2.dat};
\addplot[black,dashed,thin] coordinates {(-1,-1) (-1,-1)}; \addlegendentry{orig (BPE)}
\addplot[blue!70!black,dashed,thin]   coordinates {(-0.1,0.778) (7.3,0.778)};
\addplot[red!70!black,dashed,thin]    coordinates {(-0.1,1.022) (7.3,1.022)};
\addplot[green!50!black,dashed,thin]  coordinates {(-0.1,0.587) (7.3,0.587)};
\addplot[orange!90!black,dashed,thin] coordinates {(-0.1,0.205) (7.3,0.205)};
\addplot[black,dashed,thin,forget plot] coordinates {(1.143,0.15) (1.143,5)};
\node[font=\footnotesize,anchor=south west] at (axis cs:1.2,3.2) {M1 $\rightarrow$ M2};
\end{axis}
\end{tikzpicture}
\caption{Held-out bpb on the four domains against training bytes (solid: LD3, bf16 readings from the training log; dashed: the original model, token log-probabilities converted to bits per byte). English and code return to the original level within the first 100 steps of M1; Chinese catches up and ends slightly below the original only at the end of M2. The vertical dashed line is the M1 / M2 boundary (1.143 GB). The endpoint differs from the FP8 serving readings in Table 2 by $\leq 0.002$.}
\label{fig:bpb}
\end{figure}

\begin{table}[htbp]
\centering
\small
\caption{Held-out window set bpb (4 domains $\times$ 96 windows $\times$ 4,096 bytes; $\pm$ is the standard error across windows; orig is the sum of token logprobs divided by the byte count)}
\label{tab:bpb}
\begin{tabular}{lrrr}
\toprule
Domain & orig (BPE reading) & \textbf{LD3-base} & $\Delta$ \\
\midrule
zh (Chinese web text) & 0.778 $\pm$ 0.021 & 0.742 $\pm$ 0.018 & $-$0.036 \\
zh\_ood (Chinese, out-of-domain) & 1.022 $\pm$ 0.013 & 0.997 $\pm$ 0.013 & $-$0.025 \\
en & 0.587 $\pm$ 0.017 & 0.549 $\pm$ 0.014 & $-$0.038 \\
code & 0.205 $\pm$ 0.008 & 0.190 $\pm$ 0.008 & $-$0.015 \\
\bottomrule
\end{tabular}
\end{table}
\subsection{Chinese}
Chinese is what byteification is most likely to hurt, so two small probe sets were used to watch it between M1 / M2, and a full-character recitation was added at the end.

\begin{itemize}
\item \textbf{Five-way terminology probe}. 12 Chinese–English terms (gradient descent, backpropagation, hash table, deadlock, etc.), each asked in 5 ways: A English $\rightarrow$ Chinese, answer with the term only; B Chinese $\rightarrow$ English; C asked in Chinese, answered in Chinese; D asked in English with a Chinese answer required; E asked in Chinese with the English given in parentheses; greedy, with thinking, cap 8,192 bytes, answer section judged manually. At the end of M1: A 0/12, B 6/12, C 6/12, D 11/12, E 12/12: the knowledge is there and the English pathway works; what is broken is the link between "Chinese word form $\leftrightarrow$ concept", which is direct evidence for Section 3.3's "Chinese is the new skill". By M2 step 3,000, A is already 12/12 and B 10/12, unchanged thereafter.
\item \textbf{30-question greedy probe}. English 10 / Chinese 5 / code 5 / format 6 / gradient 4, temperature 0, cap 16,384 bytes, only the answer section after \texttt{\textless{}/think\textgreater{}} is judged, and code questions are actually run against test cases. M1 29/30 $\rightarrow$ M2@3000 28 $\rightarrow$ M2@6000 30 $\rightarrow$ LD3-base 30/30. Both probe sets are used only for decisions during training; n is too small for them to serve as main results.
\item \textbf{Full-character recitation of 12,126 characters} (E11). The character list shares its source with the character dictionary used in training (8,105 standard characters + 4,021 traditional characters); in groups of 20 (607 groups), the prompt asks the model to "recite exactly, without missing a character", greedy, non-thinking; the output is aligned character by character with the input, recording missing / wrong characters and invalid bytes; both columns on the same GPU under the same protocol, each run twice (dictionary order, then shuffled and regrouped). Over the two passes (24,252 character slots), orig gets 3 characters wrong (汨 $\rightarrow$ 汩 and two rare characters copied as look-alikes) and LD3-base 1 (欽 copied as a look-alike that appeared earlier in the same group); all are copies of a neighboring or similar-looking character, none is missing; both columns have zero invalid UTF-8 and zero cap hits. Including the 4,021 traditional characters and rare characters from the extension blocks, the byte model writes characters byte by byte no worse than the original model; the difference is within noise.
\end{itemize}
\subsection{Character-Level Tasks: What Bytes Buy (E5)}
The question to answer: a BPE model cannot see spelling, whereas a byte model sees it by construction—how large is this theoretical advantage at 27B, and on Chinese? Table 3 gives the macro-average over the four suites (all items in non-thinking mode; thinking mode only supplements CUTE and our suite with 100 items per task), and Table 4 gives per-task results.

\begin{table}[htbp]
\centering
\footnotesize
\caption{Character-level tasks, macro-average (exact match \%; all items in non-thinking mode, 100 items per task in thinking mode)}
\label{tab:e5macro}
\begin{adjustbox}{max width=\textwidth}
\setlength{\tabcolsep}{3.5pt}
\begin{tabular}{llrlrrl}
\toprule
Suite & \#tasks & orig non-thinking & \textbf{LD3-base non-thinking} & $\Delta$ & orig thinking & \textbf{LD3-base thinking} \\
\midrule
CUTE (English) & 14 & 75.2 & \textbf{84.0} & +8.8 & 89.4 & \textbf{91.2} \\
EXECUTE-eng & 12 & 76.0 & \textbf{83.2} & +7.2 & — & — \\
EXECUTE-zho & 12 (excl. extra\_*) & 81.6 & \textbf{93.9} & +12.2 & — & — \\
Our Chinese character suite & 15 & 54.4 & \textbf{64.0} & +9.7 & 87.8 & \textbf{89.1} \\
\bottomrule
\end{tabular}
\end{adjustbox}
\end{table}
Reading of the non-thinking mode (temperature 0, cap 1,024 bytes / 256 tokens, zero errors in both columns):

\begin{itemize}
\item \textbf{Character operations lead across the board, with larger margins on Chinese than on English.} On deleting / inserting / substituting / swapping characters, orthographic variants (orth) and finding characters, LD3-base leads by +10 to +43: CUTE sub\_char 37.2 $\rightarrow$ 71.2, ins\_char 21.2 $\rightarrow$ 44.2; EXECUTE-zho swap\_char 48.5 $\rightarrow$ 91.5, ins\_char 65.5 $\rightarrow$ 99.5; our suite zh\_swap\_char 13.0 $\rightarrow$ 50.5, zh\_del\_char 62.0 $\rightarrow$ 94.5, zh\_sub\_char 53.0 $\rightarrow$ 85.5. Word-level operations (del\_word / ins\_word / sub\_word / contains\_word) are already near-perfect in both columns and stay level.
\item \textbf{Not achieved by writing more.} The byte model's output byte count is about 2–4$\times$ the original's token count, consistent with the sequence-length ratio; cap hits are close to 0 in both columns. The original only gets these items right in thinking mode (CUTE 75.2 $\rightarrow$ 89.4, our suite 54.4 $\rightarrow$ 87.8), at a cost of hundreds to thousands of tokens per item (CUTE del\_word averages 3,306 tokens, 36/100 cap hits).
\item \textbf{The byte model drops on Chinese character knowledge tasks.} zh\_homophone (pick the homophone among four choices) 96.5 $\rightarrow$ 57.5, zh\_strokes 63.0 $\rightarrow$ 47.0, zh\_pinyin 98.0 $\rightarrow$ 91.5; zh\_radical +5.0 and traditional/simplified conversion +1.9 are level. Giving the pinyin of a single character is still acceptable, but comparing the pronunciations of four characters clearly is not: of the 85 wrong answers, 71 chose A (defaulting to the first option when unsure). Pronunciation and stroke count are "character $\rightarrow$ attribute" lookup knowledge; after byteification the 3 bytes of each Chinese character must be relearned, and this knowledge has not fully come back—the same direction as C-Eval $-$4.5 in Section 5.1. This is the point this section most needs to state honestly: \textbf{byteification buys operational ability at the cost of part of the character-level knowledge.}
\item \textbf{Counting and positioning are weak in both columns.} On zh\_count / zh\_nth / zh\_index / zh\_reverse the original scores 6–23 and the byte model 25–37; under non-thinking greedy decoding neither 27B model counts positions reliably, and byteification merely errs a little less.
\item Excluded tasks: EXECUTE-zho extra\_c2g / extra\_contains / extra\_g2c carry no task instructions (Section 4.4); orig 63.5 / 0 / 0, LD3-base 15.5 / 0 / 0, not included in the tables.
\end{itemize}
Table 4, per-task results (non-thinking mode; n: CUTE 400, EXECUTE 200, ours 200, traditional/simplified 107):

\begin{table}[htbp]
\centering
\footnotesize
\caption{Character-level tasks, per task (non-thinking mode, exact match \%; n: CUTE 400, EXECUTE 200, ours 200, traditional/simplified conversion 107)}
\label{tab:e5task}
\begin{adjustbox}{max width=\textwidth}
\setlength{\tabcolsep}{3.5pt}
\begin{tabular}{llrrrllrrr}
\toprule
Suite & Task & orig & \textbf{LD3-base} & $\Delta$ & Suite & Task & orig & \textbf{LD3-base} & $\Delta$ \\
\midrule
CUTE & contains\_char & 83.8 & 94.0 & +10.2 & EXECUTE-zho & contains\_char & 99.5 & 100.0 & +0.5 \\
CUTE & contains\_word & 100.0 & 99.8 & $-$0.2 & EXECUTE-zho & contains\_word & 99.5 & 100.0 & +0.5 \\
CUTE & del\_char & 67.5 & 82.8 & +15.2 & EXECUTE-zho & del\_char & 96.0 & 99.0 & +3.0 \\
CUTE & del\_word & 96.8 & 96.2 & $-$0.5 & EXECUTE-zho & del\_word & 91.0 & 94.5 & +3.5 \\
CUTE & ins\_char & 21.2 & 44.2 & +23.0 & EXECUTE-zho & ins\_char & 65.5 & 99.5 & +34.0 \\
CUTE & ins\_word & 98.0 & 98.2 & +0.2 & EXECUTE-zho & ins\_word & 56.0 & 79.0 & +23.0 \\
CUTE & orth & 68.0 & 84.5 & +16.5 & EXECUTE-zho & spell & 100.0 & 100.0 & 0.0 \\
CUTE & sem & 95.8 & 94.2 & $-$1.5 & EXECUTE-zho & spell\_inverse & 100.0 & 100.0 & 0.0 \\
CUTE & spell & 100.0 & 100.0 & 0.0 & EXECUTE-zho & sub\_char & 83.5 & 93.0 & +9.5 \\
CUTE & spell\_inverse & 100.0 & 100.0 & 0.0 & EXECUTE-zho & sub\_word & 91.5 & 98.0 & +6.5 \\
CUTE & sub\_char & 37.2 & 71.2 & +34.0 & EXECUTE-zho & swap\_char & 48.5 & 91.5 & +43.0 \\
CUTE & sub\_word & 96.5 & 95.5 & $-$1.0 & EXECUTE-zho & swap\_word & 48.5 & 72.0 & +23.5 \\
CUTE & swap\_char & 11.0 & 22.5 & +11.5 & ours & zh\_count & 12.0 & 36.5 & +24.5 \\
CUTE & swap\_word & 76.8 & 92.8 & +16.0 & ours & zh\_reverse & 6.0 & 25.5 & +19.5 \\
EXECUTE-eng & contains\_char & 94.0 & 96.0 & +2.0 & ours & zh\_nth & 23.0 & 24.5 & +1.5 \\
EXECUTE-eng & contains\_word & 100.0 & 100.0 & 0.0 & ours & zh\_index & 22.0 & 35.0 & +13.0 \\
EXECUTE-eng & del\_char & 73.5 & 88.5 & +15.0 & ours & zh\_contains & 81.0 & 90.5 & +9.5 \\
EXECUTE-eng & del\_word & 97.5 & 96.0 & $-$1.5 & ours & zh\_del\_char & 62.0 & 94.5 & +32.5 \\
EXECUTE-eng & ins\_char & 11.5 & 40.5 & +29.0 & ours & zh\_ins\_char & 40.5 & 68.0 & +27.5 \\
EXECUTE-eng & ins\_word & 97.5 & 98.5 & +1.0 & ours & zh\_sub\_char & 53.0 & 85.5 & +32.5 \\
EXECUTE-eng & spell & 99.5 & 100.0 & +0.5 & ours & zh\_swap\_char & 13.0 & 50.5 & +37.5 \\
EXECUTE-eng & spell\_inverse & 99.5 & 99.5 & 0.0 & ours & zh\_s2t & 77.6 & 79.4 & +1.9 \\
EXECUTE-eng & sub\_char & 42.5 & 67.0 & +24.5 & ours & zh\_t2s & 96.3 & 98.1 & +1.9 \\
EXECUTE-eng & sub\_word & 99.5 & 99.0 & $-$0.5 & ours & zh\_pinyin & 98.0 & 91.5 & $-$6.5 \\
EXECUTE-eng & swap\_char & 11.5 & 25.0 & +13.5 & ours & zh\_homophone & 96.5 & 57.5 & $-$39.0 \\
EXECUTE-eng & swap\_word & 85.5 & 88.5 & +3.0 & ours & zh\_strokes & 63.0 & 47.0 & $-$16.0 \\
 &  &  &  &  & ours & zh\_radical & 71.5 & 76.5 & +5.0 \\
\bottomrule
\end{tabular}
\end{adjustbox}
\end{table}
In thinking mode (100 items per task, temperature 0.6 / top\_p 0.95, cap 16,384 bytes / 4,096 tokens) both columns are near-perfect and the gap narrows to CUTE +1.9 and our suite +1.3, but the per-task breakdown still has structure:

\begin{itemize}
\item \textbf{The byte model is still higher on character operations}, by +4 to +34: zh\_reverse 59 $\rightarrow$ 93, swap\_word 77 $\rightarrow$ 96, zh\_swap\_char 84 $\rightarrow$ 98, ins\_char 85 $\rightarrow$ 95; on the four Chinese counting / positioning tasks the byte model scores 100 / 100 / 100 / 93 (original 93 / 96 / 95 / 59).
\item \textbf{Thinking does not recover the Chinese character knowledge tasks}: stroke count 34 vs 69, pinyin 88 vs 99, homophone 88 vs 99, radical 64 vs 72—lookup knowledge is not something reasoning can make up for.
\item \textbf{Two tasks where the byte model instead drops}: CUTE del\_word 64 $\rightarrow$ 56 and orth 67 $\rightarrow$ 48, caused by cap hits (43/100 and 53/100, original 36 / 33), with an average output of about 10,000 bytes. On a supplementary re-sampling run the cap-hit count rose further, indicating a probabilistic case of the thought never closing rather than a few specific items. This is direct evidence for the byte-model-specific problems of Section 6, and the reason the thinking-mode conclusions are less clean than the non-thinking ones.
\end{itemize}
Per-task thinking-mode results are in Appendix Table A1. Figure 2 plots the $\Delta$ of Table 4 per task, with operation and knowledge tasks in different colors.

\begin{figure}[htbp]
\centering
\begin{tikzpicture}
\begin{axis}[
  width=0.95\textwidth, height=7.2cm,
  ybar, bar width=4.2pt,
  ylabel={$\Delta$ exact match (LD3-base $-$ orig, points)},
  ymin=-45, ymax=48,
  xtick=data, xticklabels from table={figs/delta.dat}{label},
  x tick label style={rotate=90,anchor=east,font=\scriptsize,yshift=-2pt},
  ytick={-40,-20,0,20,40},
  grid=major, grid style={gray!25},
  legend style={at={(0.02,0.97)},anchor=north west,font=\footnotesize,cells={anchor=west}},
  tick label style={font=\footnotesize}, label style={font=\small},
  enlarge x limits=0.015,
  axis on top,
]
\addplot[fill=blue!60!black,draw=none]   table[x=idx,y=op]   {figs/delta.dat}; \addlegendentry{character / word operations}
\addplot[fill=red!70!black,draw=none]    table[x=idx,y=know] {figs/delta.dat}; \addlegendentry{character knowledge (pinyin / strokes / radical / trad.--simp.)}
\addplot[fill=gray!60,draw=none]         table[x=idx,y=cnt]  {figs/delta.dat}; \addlegendentry{counting / positioning}
\addplot[black,thin,forget plot] coordinates {(-1,0) (60,0)};
\end{axis}
\end{tikzpicture}
\caption{Per-task $\Delta$ on the character-level suites (non-thinking mode; the data of Table 4). From left: CUTE (14 tasks), EXECUTE-eng (12), EXECUTE-zho (12), our Chinese suite (15). Operation tasks are almost all positive, with larger gains on Chinese than on English; character-knowledge tasks (homophone, strokes, pinyin) are negative; on counting / positioning both models are weak and the byte model errs slightly less.}
\label{fig:delta}
\end{figure}
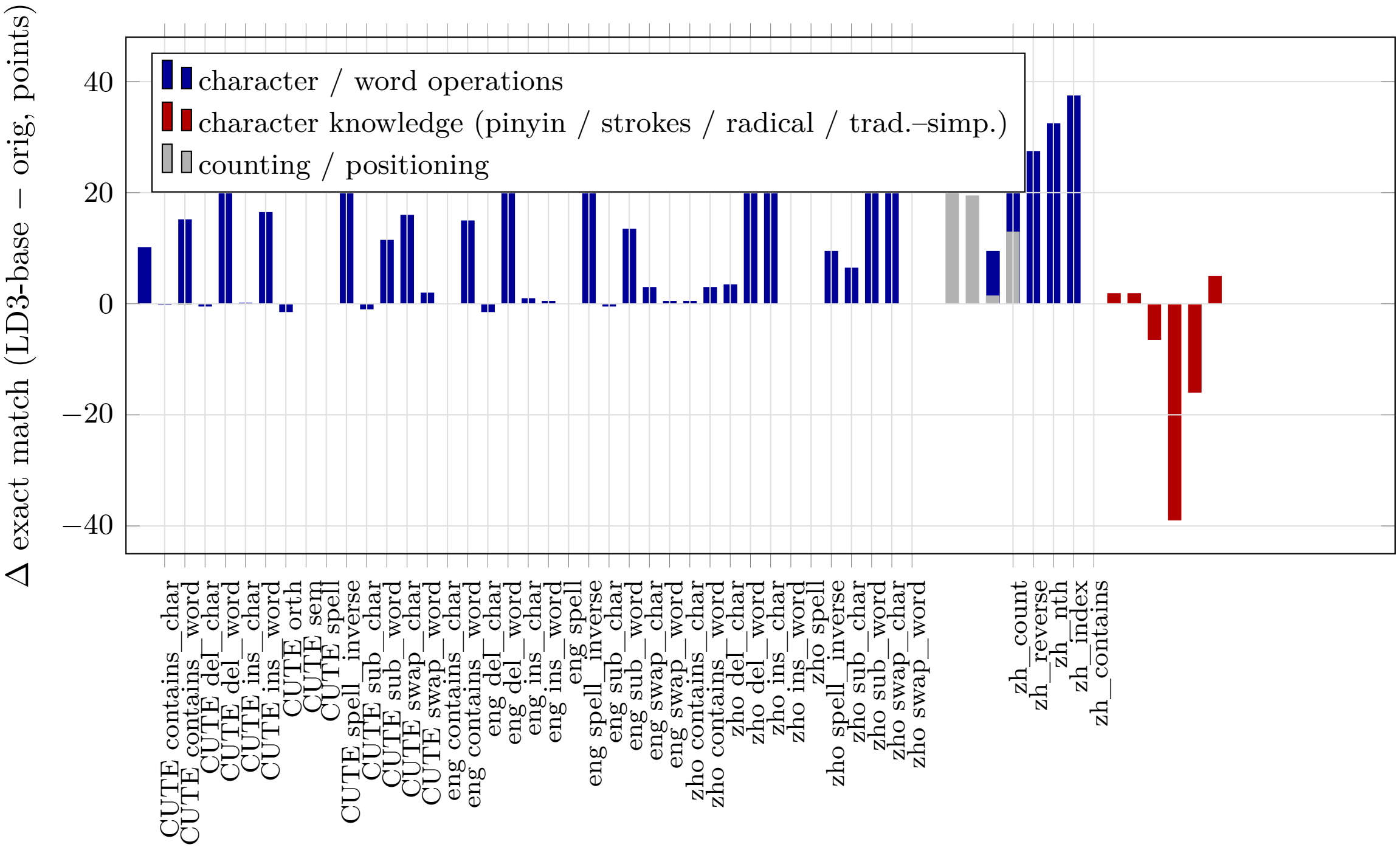

\subsection{Long Context}
LD3-base's training windows are only 64K (M1) and 16K (M2) bytes; the positional encoding was left untouched and no dedicated long-position training was done. This section asks how much of the original's long-context ability remains after byteification. All four items are measured on LD3-base, \textbf{without an original-model comparison} (on the same suites the original left only a record of "no drop up to 1M", with no scores), so they demonstrate retention only and make no comparison.

\begin{itemize}
\item \textbf{Held-out bpb at 128k: 0.257} (0–8k bucket 0.196, 64–128k bucket 0.268): 95 held-out long sessions, one forward pass over the full 128k-byte window, per-position CE computed only on assistant segments, bucketed by target position. This window set is not the same as the four-domain windows of Section 5.2, and the two are not comparable.
\item \textbf{Needle test 8k–128k, 4/4 throughout}: our own documents are concatenated into a haystack, a one-line "project id $\rightarrow$ passphrase" memo is inserted at depth 0.1 / 0.5 / 0.9, the question is asked at the end, greedy decoding, hit = the passphrase appears.
\item \textbf{Ordering probe 16k–128k, 6/6 throughout}: two memos placed 0.3 / 0.6 apart, asked "which one appears first".
\item \textbf{Up to 1M bytes}: our own 8 task types (needle, verbatim transcription, hex / base64 / URL encoding and decoding, two-hop, summation, ordering, long-document reading), 46 items per length tier, with the evidence at the beginning and the question at the end, tiers differing only in padding length; greedy decoding with thinking, cap 16,384 bytes, only the answer segment is scored. With static YaRN $\times$4.5 enabled, LD3-base scores 256K 39, 512K 35, 768K 37, 1M 33/46; without YaRN, 1M 27/46; the errors are mostly in the encoding/decoding types.
\end{itemize}
The conclusion is one sentence: at positions 2–64$\times$ beyond the training window the model can still read and locate, the original's positional ability survives byteification, and length extrapolation can simply reuse the original's YaRN. The LongBench-v2 row of E1 is dominated by the output cap and is not used as long-context evidence (Section 5.1 footnote).

\subsection{Efficiency: What Bytes Cost (E6)}
Sequences are about 3–4$\times$ longer; the cost lies not in slower computation per step but in less output per step. Table 5 reports measurements on the same RTX 6000 Ada 48G with the same serving stack (single-stream: mean over 10 prompts per language; concurrency 16; prefill: median of 3 runs; protocol of Section 4.5).

\begin{table}[htbp]
\centering
\footnotesize
\caption{Efficiency on the same GPU (RTX 6000 Ada 48G, sglang, FP8; single-stream is the mean over 10 prompts per language, concurrency 16, prefill is the median time to first token over 3 runs)}
\label{tab:eff}
\begin{adjustbox}{max width=\textwidth}
\setlength{\tabcolsep}{3.5pt}
\begin{tabular}{llllllll}
\toprule
Config & single-stream token/s & single-stream Chinese chars/s & single-stream English chars/s & conc. 16 Chinese chars/s & conc. 16 English chars/s & TTFT 16,384 B & TTFT 65,536 B \\
\midrule
orig & 22.9 & 33.1 & 115.4 & 482 & 1,441 & 1.16 s & 4.52 s \\
\textbf{LD3-base} & 24.3 & \textbf{7.7} & \textbf{24.3} & \textbf{97} & \textbf{332} & \textbf{5.83 s} & \textbf{29.57 s} \\
ratio (LD3-base / orig) & 1.06 & 0.23 & 0.21 & 0.20 & 0.23 & 5.0$\times$ & 6.5$\times$ \\
orig + NEXTN draft & does not start on a single GPU &  &  &  &  &  &  \\
LD3-base + NEXTN draft K=8 & does not start on a single GPU &  &  &  &  &  &  \\
\bottomrule
\end{tabular}
\end{adjustbox}
\end{table}
Reading:

\begin{itemize}
\item \textbf{Identical per-step speed.} Single-stream token/s is nearly the same in both columns (22.9 vs 24.3), a direct reflection of the shared backbone; aggregate token/s at concurrency 16 is also close (329 vs 307). The entire difference is in output per step: for the original, 1 token $\approx$ 1.45 Chinese characters or 5.0 English characters; for the byte model, 1 step = 1 byte $\approx$ 0.32 Chinese characters or 1 English character. Hence both Chinese characters per second and English characters per second fall to 1/4–1/5 of the original, and the ratio is unchanged under concurrency.
\item \textbf{Prefill loses more.} For the same 65,536-byte Chinese input, the original has about 20,000 tokens and the byte model 65,622 tokens; time to first token goes 4.52 $\rightarrow$ 29.57 s (6.5$\times$), twice the token-count ratio (about 3.2). The excess is an engineering gap: the byte model runs through add-on chunked GDN and attention kernels that lack the tuning of the BPE path. This is not a property of the model itself, but it is the current state.
\item \textbf{The draft head does not fit on a single 48G GPU.} Neither NEXTN configuration starts (with the 27B FP8 weights plus the draft head, the KV pool no longer fits), so Table 5 has no draft rows. The evidence that MTP is especially effective for the byte model can only be cited from records on other hardware: with K = 8, about 5.7 bytes accepted per round, about 3.3$\times$ single-stream and about 1.8$\times$ at concurrency 16. Scaled accordingly, single-stream Chinese characters per second would return to 0.7–0.8$\times$ the original and to about 0.4$\times$ under concurrency—\textbf{even with the draft head, throughput on the same hardware remains below the original.}
\item One more number under the end-to-end protocol: the wall-clock time for all E1 benchmarks in both modes on the same GPU is orig 7 h 44 min and LD3-base 35 h 57 min (4.6$\times$), consistent with the individual ratios in Table 5.
\end{itemize}
\subsection{Multilingual (E11)}
The continued training of LD3-base used Chinese and English data only (Section 3.2). This section uses a smoke test of 22 languages $\times$ 3 items to see how much of the "byte channel" and of the generation ability remain, both columns on the same GPU under the same protocol (E11): (1) recite a sentence in the language verbatim (character-by-character comparison after NFC normalization); (2) translate the same Chinese sentence into the language; (3) ask a capital-city common-knowledge question in the language and require an answer in that language; greedy, one run without thinking (caps 512 / 512 / 1,024) and one with thinking (4,096 / 4,096 / 8,192), with the same max\_tokens values for both columns (for the BPE model this amounts to about 3–4$\times$ the text budget, so the original model's numbers are an upper bound). Translation and QA are scored automatically: the fraction of letters in the output that fall in the target script (script\_match), 8-gram repetition rate and cap hits to detect loops, and whether the QA answer contains the gold string; kept = translation and answer both have script\_match $\geq$ 0.9 and do not loop, degraded = either item loops or has script\_match < 0.5, marginal = the rest.

\begin{table}[htbp]
\centering
\small
\caption{22-language smoke test (E11; recitation = number of languages recited exactly, QA = number answered correctly, kept / marginal / degraded = number of languages per automatic tier by script and looping)}
\label{tab:ml}
\begin{tabular}{lrrrrr}
\toprule
Model / mode & Recitation & QA correct & Kept & Marginal & Degraded \\
\midrule
orig non-thinking & 22/22 & 22/22 & 21 & 1 & 0 \\
orig thinking & 21/22 & 22/22 & 22 & 0 & 0 \\
\textbf{LD3-base} non-thinking & 22/22 & 13/22 & 11 & 0 & 11 \\
\textbf{LD3-base} thinking & 22/22 & 14/22 & 13 & 0 & 9 \\
\bottomrule
\end{tabular}
\end{table}
Reading:

\begin{itemize}
\item \textbf{The byte channel is intact; generation splits by script.} Recitation is nearly perfect in both columns (in thinking mode the original model repeats tone marks in one Thai sentence, a BPE token-boundary artifact on Thai; the byte model is 22/22). In generation the original model keeps all 22 languages; the byte model keeps only the major Latin- and Cyrillic-script languages plus Japanese (11 in non-thinking mode: Russian, Japanese, Arabic, German, French, Spanish, Portuguese, Vietnamese, Turkish, Polish, Italian), while non-Latin scripts such as Greek, Korean, Persian, the Indic languages, Thai and Hebrew degrade: translations loop or become word salad, and QA answers miss the question (in Hindi it says the capital of India is "Hindi").
\item \textbf{The degradation is a data problem, not an architectural one.} Intact copying shows the byte channel is not broken; what is missing is supervised bytes in these languages, as the two-stage curriculum contains only Chinese and English (Section 3.2). Bringing them back would require mixing the corresponding languages into the continued-training data, which this paper did not do.
\item \textbf{Thinking is double-edged for low-resource scripts.} With thinking on, Greek, Korean, Ukrainian and Indonesian move from degraded back to kept (the translation is written out in one go); Vietnamese, Swahili and Arabic drop instead, in every case because the thinking itself loops in that language up to the cap and the body is empty; the net effect is +2 languages (11 $\rightarrow$ 13), at 1,356 s against 171 s for the original model. This is the same phenomenon as "the thought never closes" in Sections 5.4 and 6.1.
\end{itemize}
\section{Failure Modes Specific to Byte Models}
Spreading one BPE token over about 4 bytes amounts to splitting every sampling decision into 4. The four problems below all stem from this; they do not exist, or are not serious, in BPE models, and must be handled separately in byte models.

\subsection{Runaway under Greedy Decoding}
\textbf{Phenomenon.} At temperature 0 the byte model may write all the way to the output cap without closing. In one non-thinking GPQA-Diamond greedy evaluation on the pilot lines, 34 of 198 items (17\%) filled the 16,384-byte cap with no answer line; the BPE original built on the same weights had 0 items with no extractable answer under greedy decoding, with an output p90 of only 2,005 characters. In the thinking-mode greedy evaluation of the first-round 27B byteification, 42\% of IFEval samples hit the 32,768-byte cap. On LD3-base the same phenomenon appears in the thinking-mode character-level tasks: in CUTE del\_word and orth, 43/100 and 53/100 samples respectively hit the 16,384-byte cap; whereas on the seven benchmarks of the main table in Section 5, the number of LD3-base thinking-mode cap hits is no higher than the original's on every item, so this is a problem of particular task types, not a general "thinking that never stops".

\textbf{Cause.} Runaway samples split into two kinds: half are byte-level infinite loops (\texttt{\textbackslash{}end\{bmatrix\}} repeated endlessly, consecutive newlines, 60-character-block repetition rate $\geq$ 0.9), and half are "Let me reconsider… Actually…" thoughts that never finish (repetition rate $\approx$ 0). The former is the greedy chain locking into a cycle more easily at the byte level: each Chinese character takes 3 steps, each step's argmax is near-deterministic, and once it enters a high-probability short cycle there is no randomness to bring it out. The latter is long-range closing never having been supervised: end symbols are all counted in the loss, but no training window ends with a loss-bearing end symbol, so once thinking exceeds the p99 length the model has never seen "time to stop"—this is exposure bias, essentially a problem of the RL domain.

\textbf{Fix.} Three measures that leave the model untouched: always sample rather than decode greedily in evaluation and serving; align the output cap with BPE models at bytes ÷ 4 (Section 4.3); report the counts of "no extractable answer / cap hits / unjudgeable" alongside scores, because "counting no extractable answer as wrong" turns all such samples into inability. To these is added the decoding guard of Section 3.6 (budget + circuit breaker). The measure that changes the model is the self-sampling RFT of Section 3.4.

\textbf{Evidence.} After the pilot lines switched from greedy to sampling, GPQA items with no extractable answer fell 34 $\rightarrow$ 24 and the score rose 54.0 $\rightarrow$ 60.6; adding the guard, thinking mode went 48.0 $\rightarrow$ 60.6. But the number of correct answers for all three guard versions stopped at 120/198 (the original BPE model: 170): the decoding side can only turn "unfinished" into "finished"; whether the finished output is correct depends on the model.

\subsection{Byte-wise Sampling Breaks UTF-8}
\textbf{Phenomenon.} With sampling on, symbols that should not appear occasionally show up in the output: in samples the multiplication sign \texttt{×} became a geometric-shape symbol, and a superscript symbol became a currency symbol (both are valid characters sharing the same lead byte as the original character but with a different continuation byte). The worse case is producing an invalid byte sequence outright, which the decoding end can only display as the replacement character.

\textbf{Cause.} The three bytes of a Chinese character or symbol are one sampling decision in a BPE model and three in a byte model. After the lead byte, at temperature 0.7 there is still a small probability of sampling another valid continuation byte (valid but wrong) or a non-continuation byte (invalid sequence). The former is an inherent cost of byte sampling; the latter can be eliminated on the decoding side.

\textbf{Fix.} Add a UTF-8 validity mask on the decoding side (Section 3.6). The mask depends on the state "the byte just sampled", so the inference framework's overlap scheduling must be turned off: otherwise the output the guard reads lags by one token, and when a lead byte has just been emitted the guard still believes it is at a character boundary and instead bans all valid continuation bytes, turning the output into "lead, lead, continuation" triples. The "valid but wrong" garbling can only be handled by tightening sampling; the mask cannot cure it.

\textbf{Evidence.} This bug was found on a pilot-line checkpoint: for the same Chinese prompt, with the guard on + overlap scheduling the output was almost entirely replacement characters without a single complete Chinese character; with the guard fully off there were occasional invalid sequences; with the guard on + overlap scheduling off the output was entirely valid. It once led us to believe the model had not learned Chinese encoding; all historical scores obtained with the guard were therefore re-examined.

\subsection{Presence / Frequency Penalties Are Ineffective on a 256-Entry Vocabulary}
\textbf{Phenomenon.} The common remedy for repetition is to turn on presence / frequency penalty. After deployment, LD3-base reproduced repetition on real agent inputs (16,000-token prefix), and adding a frequency / presence penalty had no effect.

\textbf{Cause.} Both penalties keep accounts per token id, lowering ids that have already appeared. In a 150,000-entry vocabulary "has appeared" is informative; among 256 bytes every common byte has appeared within a few steps, so the penalty amounts to shifting all bytes together and cannot suppress n-gram-level loops. Repetition suppression for byte models must be done at the n-gram level, for which inference frameworks have no ready-made implementation.

\textbf{Fix.} On the decoding side, use the n-gram-level loop detection of Section 3.6, but feed it only to the circuit breaker and leave the logits untouched otherwise—byte-wise repetition banning would falsely trigger on 41\% of decoding steps in normal text. On the training side, use the on-policy RFT / DPO of Section 3.4.

\textbf{Evidence.} See the guard results in Section 6.1 and the training results in Section 6.4; the penalties themselves have no positive number to report.

\subsection{Repetition and Its Treatment}
\textbf{Phenomenon.} 46\% of LD3-base's self-generated answers contain 8-gram repetition; on the same prompts another commercial model has 6\% and another open-source model 13\%. Chinese thinking repetition under long context (24–32K prefix) is a probabilistic event: the original Qwen3.8-27B on the same long Chinese task also repeated 2 times out of 10 samples (20\%), and 0/3 when sampled 3 times. Under a stricter criterion (loops / whole-sentence repetition) the rate is only 4–6\%; the two numbers are not directly comparable.

\textbf{Cause.} Three layers. First, the byte level amplifies exposure bias by about 4$\times$, and once in a short cycle there is no randomness to bring it out (Section 6.1). Second, the original model already has some probability of repeating itself while thinking; after byteification Chinese supervised bytes are fewer (see Section 3.2), so the repetition rate of Chinese thinking is higher. Third, part of the "stutter" is not the model's illness: the "你你好" and "感感文件" seen in client-side streaming are caused by the inference framework's incremental detokenization computing offsets by character count and re-sending text when the three steps of one Chinese character arrive on a byte vocabulary; the non-streaming output of the same request is clean, and the stutter rate is only 0.2–0.5\%.

\textbf{Fix.} Prompting is ineffective (LD3-base ignores the style guide written into the system prompt), penalties are ineffective (Section 6.3), so it can only be trained in: Section 3.4's self-sampling (guard off) $\rightarrow$ RFT $\rightarrow$ length-normalized DPO. The pairing rule determines what DPO learns: between chosen and rejected there must be no systematic difference other than the "illness" (length, amount of thinking, thinking language). Verifying repetition requires $\geq$ 10 samples and reporting the trigger rate.

\textbf{Evidence.} On the M3 chain the bad-sample rate (repetition / stutter / cap hits) went 11.9\% $\rightarrow$ 7.7\% (direct DPO) $\rightarrow$ 0.2\% (RFT + two rounds of DPO); the cost was precise reasoning: GSM8K 300-item greedy dialogue 97.7\% $\rightarrow$ 80.7\%, and RFT before DPO recovered it to 92.7\%. But the 0.2\% must be discounted: in the last round, 211 of 252 thinking answers had thoughts shorter than 60 characters, median 28 characters; most of it comes from "not thinking". Long-context repetition has its own conclusions: SFT works (200 clean long prefixes brought the trigger rate on the long Chinese task from 20\% $\rightarrow$ 0\%; 513 Chinese 32K items brought repetition on English agent inputs from 2/10 $\rightarrow$ 1/10), while DPO with too few pairs does not (169 pairs and 218 pairs, one run each). Up to LD3-base, repetition has not been cured; this is one of the limitations in Section 7.

\section{Limitations}
\begin{itemize}
\item \textbf{Single seed, single run.} The decoding randomness of thinking-mode sampling at 0.6 was not measured; the bootstrap intervals cover only item-sampling variance.
\item \textbf{The bpb comparison shares the distribution of the training data} (Section 5.2): it cannot be read as stronger modeling ability.
\item \textbf{The attribution of the non-thinking IFEval gap is unclear}: LD3-base has had no instruction-type post-training whereas the original has; this paper has no experiment isolating this factor.
\item \textbf{Efficiency is a real cost}: Chinese characters per second on the same GPU is about 1/4 of the original, and long-input prefill is 5–6.5$\times$ slower; the draft head could not be measured on a single GPU, and only data from elsewhere can be cited.
\item \textbf{Character-level knowledge is not fully recovered}: lookup knowledge of the "character $\rightarrow$ attribute" kind, such as pronunciation / stroke count, drops after byteification, in the same direction as the C-Eval loss; this paper does not answer how much Chinese data would be needed to recover it.
\item \textbf{Thinking mode is unstable}: without post-training, the byte model cannot find its way out of thinking on some task types (del\_word / orth in Section 5.4); the treatment is in Sections 3.4 / 6.4, but the treated model is outside the comparison scope of this paper.
\item \textbf{Multilingual degradation} (Section 5.7): generation in non-Latin / non-Cyrillic scripts degrades (11–13/22 languages kept against 21–22/22 for the original model) while recitation is intact; only 3 items per language were tested, so this is a smoke test rather than a benchmark, and using the same max\_tokens values for both columns gives the original model 3–4$\times$ the text budget.
\item \textbf{The vision tower was not re-adapted}; long-context repetition is not cured; some conclusions (the quantification of greedy runaway, the ineffectiveness of penalties) come from the pilot lines rather than LD3 itself, as noted in the text.
\item Contamination between evaluation and training sets was not systematically checked (C-Eval question stems were used as natural sentences in our Chinese character suite, but both models had already been evaluated on them in E1, and the character tasks do not test knowledge).
\end{itemize}
\section{Conclusion}
Converting a 27B subword model as is into a pure byte model is feasible: vocabulary 248,320 $\rightarrow$ 384, backbone untouched, 4.77 GB of supervised bytes, about 1,430 H200 GPU-hours, and most of the original's ability is retained—math, code and MMLU-Pro are within noise, while Chinese exams and non-thinking instruction following drop 4–6 points. What byteification buys is character-level manipulation ability that does not rely on thinking (especially pronounced for Chinese); what it costs is character throughput on the same hardware (down to 1/4–1/5 of the original) and part of the lookup knowledge of Chinese characters. Byte models also bring a set of failure modes of their own, which have corresponding engineering and training fixes. For those who want to try byteification on an existing model, our advice is threefold: use the target model's own output for dialogue-type supervision; put the dictionary (explicit character $\leftrightarrow$ byte supervision) first in the curriculum; at evaluation time, align the output cap at bytes ÷ 4 and record cap-hit counts per item.

\appendix\renewcommand{\thetable}{A\arabic{table}}\setcounter{table}{0}\renewcommand{\thefigure}{A\arabic{figure}}\setcounter{figure}{0}
\section{Supplementary Tables}
\begin{table}[htbp]
\centering
\footnotesize
\caption{Character-level tasks, per task, thinking mode (100 items per task, temperature 0.6 / top\_p 0.95, cap 16,384 bytes / 4,096 tokens; "cap hits" is the number of samples hitting the output cap)}
\label{tab:e5think}
\begin{adjustbox}{max width=\textwidth}
\setlength{\tabcolsep}{3.5pt}
\begin{tabular}{llrrrrllrrrr}
\toprule
Suite & Task & orig & \textbf{LD3-base} & $\Delta$ & cap hits orig / LD3 & Suite & Task & orig & \textbf{LD3-base} & $\Delta$ & cap hits orig / LD3 \\
\midrule
CUTE & contains\_char & 100 & 100 & 0 & 0 / 0 & ours & zh\_count & 93 & 100 & +7 & 0 / 0 \\
CUTE & contains\_word & 100 & 100 & 0 & 0 / 0 & ours & zh\_reverse & 59 & 93 & +34 & 0 / 0 \\
CUTE & del\_char & 94 & 98 & +4 & 1 / 2 & ours & zh\_nth & 96 & 100 & +4 & 0 / 0 \\
CUTE & del\_word & 64 & 56 & $-$8 & 36 / 43 & ours & zh\_index & 95 & 100 & +5 & 1 / 0 \\
CUTE & ins\_char & 85 & 95 & +10 & 2 / 0 & ours & zh\_contains & 99 & 100 & +1 & 0 / 0 \\
CUTE & ins\_word & 99 & 100 & +1 & 1 / 0 & ours & zh\_del\_char & 94 & 98 & +4 & 0 / 0 \\
CUTE & orth & 67 & 48 & $-$19 & 33 / 53 & ours & zh\_ins\_char & 92 & 97 & +5 & 0 / 0 \\
CUTE & sem & 94 & 94 & 0 & 0 / 2 & ours & zh\_sub\_char & 88 & 96 & +8 & 1 / 0 \\
CUTE & spell & 99 & 99 & 0 & 1 / 1 & ours & zh\_swap\_char & 84 & 98 & +14 & 1 / 0 \\
CUTE & spell\_inverse & 100 & 100 & 0 & 0 / 0 & ours & zh\_s2t & 80 & 80 & 0 & 0 / 0 \\
CUTE & sub\_char & 88 & 97 & +9 & 0 / 0 & ours & zh\_t2s & 98 & 100 & +2 & 1 / 0 \\
CUTE & sub\_word & 90 & 94 & +4 & 8 / 2 & ours & zh\_pinyin & 99 & 88 & $-$11 & 0 / 0 \\
CUTE & swap\_char & 94 & 100 & +6 & 1 / 0 & ours & zh\_homophone & 99 & 88 & $-$11 & 1 / 2 \\
CUTE & swap\_word & 77 & 96 & +19 & 23 / 3 & ours & zh\_strokes & 69 & 34 & $-$35 & 16 / 4 \\
 &  &  &  &  &  & ours & zh\_radical & 72 & 64 & $-$8 & 22 / 14 \\
\bottomrule
\end{tabular}
\end{adjustbox}
\end{table}
\begin{table}[htbp]
\centering
\small
\caption{Per-item bootstrap 95\% CIs for the main table (2,000 resamples; $\Delta$ is the per-item paired LD3-base $-$ orig; covers only item-sampling variance, not decoding randomness)}
\label{tab:boot}
\begin{tabular}{lllll}
\toprule
Dataset & Mode & orig & LD3-base & $\Delta$ [95\% CI] \\
\midrule
GSM8K & thinking & 97.0 [96.1, 97.9] & 93.0 [91.7, 94.4] & $-$4.0 [$-$5.3, $-$2.9] \\
MATH-500 & thinking & 94.0 [91.8, 96.0] & 95.4 [93.4, 97.0] & +1.4 [$-$0.8, +3.6] \\
HumanEval+ & thinking & 90.2 [85.4, 94.5] & 88.4 [83.5, 93.3] & $-$1.8 [$-$7.3, +3.7] \\
C-Eval & thinking & 89.9 [88.3, 91.5] & 85.4 [83.4, 87.2] & $-$4.5 [$-$6.3, $-$2.8] \\
MMLU-Pro & thinking & 80.1 [77.6, 82.5] & 78.8 [76.3, 81.4] & $-$1.3 [$-$3.3, +0.6] \\
LongBench-v2\textdagger{} & thinking & 27.4 [19.7, 35.9] & 39.3 [30.8, 47.9] & +11.9 [+3.4, +21.4] \\
GSM8K & non-thinking & 96.4 [95.4, 97.3] & 96.0 [94.9, 97.0] & $-$0.4 [$-$1.2, +0.5] \\
MATH-500 & non-thinking & 92.8 [90.4, 95.0] & 93.6 [91.4, 95.6] & +0.8 [$-$1.6, +3.2] \\
HumanEval+ & non-thinking & 92.7 [88.4, 96.3] & 90.2 [85.4, 94.5] & $-$2.5 [$-$6.7, +1.8] \\
C-Eval & non-thinking & 88.1 [86.3, 89.9] & 83.4 [81.4, 85.4] & $-$4.7 [$-$6.5, $-$3.0] \\
MMLU-Pro & non-thinking & 79.4 [76.9, 81.8] & 78.1 [75.6, 80.6] & $-$1.3 [$-$3.3, +0.7] \\
LongBench-v2\textdagger{} & non-thinking & 42.7 [34.2, 52.1] & 53.8 [44.4, 62.4] & +11.1 [+2.6, +19.7] \\
\bottomrule
\end{tabular}
\end{table}

\clearpage
\bibliographystyle{plainnat}
\bibliography{references}

\clearpage
\phantomsection\label{sec:zh}
\setcounter{section}{0}
\setcounter{table}{0}
\setcounter{figure}{0}
\setcounter{footnote}{0}
\renewcommand{\thesection}{\arabic{section}}
\renewcommand{\thetable}{\arabic{table}}
\renewcommand{\thefigure}{\arabic{figure}}
\renewcommand{\figurename}{图}
\renewcommand{\tablename}{表}
\renewcommand{\abstractname}{摘要}
\renewcommand{\refname}{参考文献}
\renewcommand{\appendixname}{附录}
\begin{center}
{\LARGE\bfseries \paperTitleZh\par}
\vspace{0.8em}
{\large \paperAuthorZh\par}
\vspace{0.4em}
2026 年 11 月
\end{center}
\begin{abstract}
字节级 BPE 把中文切得很碎(Qwen3.8 在中文维基上平均 3.84 字节 / token,64\% 的 token 是单字),而且把拼写藏在模型看不见的地方。本文问的是:一个已经训好的 27B 子词模型,能否不加切块结构、不加局部编码器,直接改成"一字节一 token"的纯字节模型,并保住大部分能力。我们从 Qwen3.8-27B(48 层 Gated DeltaNet + 16 层全注意力)出发,把词表从 248,320 收缩到 384(BPE 词表里本来就有的 256 个字节符号加特殊 id),其余权重一字不改,然后用两段课程(先学编码、再读书)续训,监督 4.77 G 字节:对话 / 行为类监督全部由原模型自己 on-policy 生成(字典、词典、一般对话、编码会话),知识阅读用人写的百科原文(占监督字节 54\%)。训练 8$\times$H200 178.6 h(约 1,430 GPU 时)。在同一张 GPU、同一服务栈的对照下,字节模型在 MATH-500、HumanEval+、MMLU-Pro 与非思考档 GSM8K 上与原版的差落在配对自助法区间内,在 C-Eval、思考档 GSM8K 与非思考档 IFEval 上掉 4–6 分。不开思考时它在字符级任务上全面领先(CUTE 75.2 $\rightarrow$ 84.0、EXECUTE-zho 81.6 $\rightarrow$ 93.9、自建中文字符题 54.4 $\rightarrow$ 64.0),但汉字的读音、笔画等查表知识下降,思考也补不回。代价是吞吐:每步速度相同、每步产出少 3–5 倍,同卡汉字 / 秒约为原版的 1/4,64K 字节 prefill 慢 6.5$\times$。我们还记录了字节模型特有的四种失效模式(贪心跑飞、采样打断 UTF-8、惩罚项对小词表无效、复读)及其解码与训练侧修法。

\end{abstract}
\noindent\small 本文为同一篇论文的中文版;英文版在前(第 1 页起)。两版实验与数字完全一致,参考文献见英文版末。
\normalsize
\section{引言}
子词分词器是训练前冻结的设计决定。对中文,Qwen3.8 的 byte-level BPE 在中文维基上平均 3.84 字节 / token,64\% 的 token 是单字(\texttt{tokenizer-eval/}),一个词常被切成几个碎片;对所有语言,拼写、字符位置、数字的各位对模型都不可见,字符级操作要靠"思考"把词拆开再做(5.4 节)。词表一旦定下,后面所有的训练与服务都绑在它上面。

最直接的替代是字节模型:1 字节 = 1 token,词表 256 加特殊符号。从零训一个字节模型太贵;已有的两条现实路线各有代价——保留切块结构(局部编码器 + 补丁,BLT、Bolmo),模型主干看到的仍是切块后的向量;或经字节接口做跨分词器蒸馏(1B–8B 上仍被判为开放问题)。本文走第三条路:\textbf{把一个已训好的 27B 子词模型,原样改成不切块的纯字节模型},只换词表与嵌入,主干一字不改。

我们的做法是:词表收缩到 384(BPE 词表里本来就有 256 个字节符号),然后用两段课程续训——先用字典、词典、对话学"字 $\leftrightarrow$ 字节"的编码,再读人写的百科原文(3 节)。对话 / 行为类监督全部由原版 Qwen3.8 自己生成(on-policy 自蒸馏,按轮独立补答、只监督本轮),知识阅读用人写原文(占监督字节 54\%)。监督 4.77 G 字节,8$\times$H200 178.6 h。

本文想回答的只有四个问题,也是四点贡献:

\begin{enumerate}
\item \textbf{保住了多少}(5.1、5.2):同卡、同服务栈、同解码口径的两列对照。非思考档 7 项里 4 项在配对自助法区间内看不出差别,C-Eval 与非思考档 IFEval 掉 4–6 分;思考档 GSM8K $-$4.0、C-Eval $-$4.5。
\item \textbf{字节带来了什么}(5.4):不开思考时字符级操作全面领先(CUTE +8.8、EXECUTE-zho +12.2、自建中文题 +9.7),中文幅度大于英文;但汉字的读音 / 笔画等查表知识掉了,思考也补不回。
\item \textbf{代价是什么}(5.6):每步速度相同,每步产出少 3–5 倍;同卡汉字 / 秒降到 1/4,64K 字节 prefill 慢 6.5$\times$;草稿头是补回的路,但 48G 单卡装不下。
\item \textbf{字节模型特有的失效模式与修法}(6 节):贪心跑飞、按字节采样打断 UTF-8、惩罚项对小词表无效、思考绕不出来;以及对应的护栏与后训练方法。
\end{enumerate}
我们不声称字节模型更强。诚实的结论是:\textbf{在 27B 规模、约 1e21 FLOPs 的续训预算下,不切块的字节化可以追平原版的大部分能力,换来字符级操作的收益,付出吞吐与一部分字级知识的代价。}评测脚本、自建中文字符题集与本文全部评测的判分 / 自助法代码公开于 \texttt{https://huggingface.co/datasets/leiwhere/ld3-byte-eval};权重与训练语料暂不发布。

\section{相关工作}
\textbf{从零训练的字节 / 字符模型。} ByT5 \citep{xue2022byt5} 把 T5 直接喂 UTF-8 字节,是"扁平字节"的早期代表;CANINE \citep{clark2022canine} 与 Charformer \citep{tay2022charformer} 用下采样或可微的软子词把字符序列压短;MEGABYTE \citep{yu2023megabyte}、SpaceByte \citep{slagle2024spacebyte}、BLT \citep{pagnoni2025blt} 与 H-Net \citep{hwang2026hnet} 把"切块"做进架构——固定长、按空格、按熵或端到端学出的边界,再在块上跑主干。MambaByte \citep{wang2024mambabyte} 与 EvaByte \citep{zheng2025evabyte} 保持扁平字节,但分别换成 SSM 主干或从零训练 6.5B。这些工作都从零训练;本文的出发点不同:一个已经花费数十万亿 token 训好的子词模型,能否不重训主干、不加切块结构,直接改成字节模型。

\textbf{把子词模型改成字节模型。} BLT 报告了用 Llama 3.1 8B 初始化全局 Transformer 的实验(220B token),保留 patching 与新训的局部编解码器 \citep{pagnoni2025blt}。Bolmo \citep{minixhofer2025bolmo} 系统地做了这件事:Olmo 3 7B 改成字节级"潜分词器语言模型",字节池化成 patch、mLSTM 局部编解码器,两阶段共约 49B token(第二阶段 39.3B);同样保留切块,主干按 patch 运行。\citet{bao2026distillbyte} 用渐进蒸馏把 token 老师的表示对齐到 H-Net 式层级字节学生($\leq$ 4B)。\citet{singh2026bld} 只把字节当作跨分词器蒸馏的接口,学生仍是子词模型,并报告 1B–8B 上一致收益"仍难以获得"。推理期方法 \citep{hayase2025bytesampling} 不训练,把任意 BPE 模型变成字节级采样器,模型内部仍按子词计算。与它们相比,本文(i)完全不切块,主干对每个字节各算一步;(ii)不引入外部老师,监督来自原模型自己的输出与人写原文;(iii)规模 27B、以中文为重点,并给出同卡的效率代价。不切块的代价是序列长度不压缩,5.6 节如实报告。

\textbf{词表迁移。} ZeTT \citep{minixhofer2024zett}、FOCUS \citep{dobler2023focus}、TokAlign \citep{li2025tokalign} 研究换一套子词 tokenizer 时如何初始化新嵌入、少步数恢复。本文是其极端情形:新词表只剩 256 个字节符号,而这 256 行本就在原嵌入表中(GPT-2 式字节级 BPE \citep{radford2019gpt2}),所以不需要任何初始化技巧,困难全部转移到续训。

\textbf{自生成数据与 on-policy。} Self-Instruct \citep{wang2023selfinstruct}、STaR \citep{zelikman2022star} 与 RFT \citep{yuan2023rft} 用模型自己(过滤后)的输出训自己;SDFT \citep{yang2024sdft} 让种子模型改写任务答案以缓解遗忘;\citet{chen2026retaining} 把 RL 少遗忘归因于 on-policy 数据。本文的语料配方(3.2 节)与后训练(3.4 节)建立在同一观察上——对话类监督必须是目标模型自己当前的输出——并把它用在词表替换这个分布剧变的设定下;DPO \citep{rafailov2023dpo} 用作复读治理的手段,带长度归一化。

\textbf{底座与基准。} Qwen3.8-27B \citep{qwen2026qwen38} 是 Gated DeltaNet \citep{yang2025gdn} 与全注意力按 3:1 混合的稠密模型(同系架构说明见 \citep{qwen2025qwen3next,qiu2026qwen38next}),原生带一层 MTP \citep{gloeckle2024mtp,deepseek2024v3}。字符级任务用 CUTE \citep{edman2024cute} 与其多语种扩展 EXECUTE \citep{edman2025execute}。

\section{方法}
\subsection{词表收缩}
Qwen3.8-27B 的分词器是 GPT-2 式的字节级 BPE(byte-level BPE):词表里本来就有 256 个单字节符号,所有多字节 token 都由它们合并而来。改成纯字节模型因此不需要新建任何嵌入,只需删掉全部合并项,保留 256 个字节行和对话模板依赖的特殊符号。收缩后的词表是 384 项:字节 0–255 按字节值排,随后是 \texttt{\textless{}\textbar{}endoftext\textbar{}\textgreater{}}=256、\texttt{\textless{}\textbar{}im\_start\textbar{}\textgreater{}}=257、\texttt{\textless{}think\textgreater{}}=280 等 35 个命名特殊符号,其余 93 个位置是占位(全表见附录 B)。tokenizer 里没有任何 merge 规则,一个 UTF-8 字节就是一个 token。

主干一字不改:64 层 = 48 层 GatedDeltaNet + 16 层全注意力(每 4 层一个),hidden 5120,\texttt{model\_type} 仍是 \texttt{qwen3\_5},现成推理框架直接装载。参数只在两端变化:嵌入与 lm\_head 各从 248,320 $\times$ 5,120 $\approx$ 1.27B 缩到 384 $\times$ 5,120 $\approx$ 2M,语言模型部分从 26.9B 变成 24.4B;视觉塔 0.46B 原样冻结。收缩脚本按 token 名字取行而不按 id 偏移;产物用特殊 id 表与对话模板的 md5 逐项核对(两处容易漏的错误:config 里的视觉 token id、新版 transformers 对特殊符号的重排)。

代价是序列变长:原分词器在中文上 3.84 字节/token、英文 4.35,字节版序列约为原来的 4 倍,16K 字节窗大致相当于 BPE 的 4k token。中文文本里只有 15.9\%(数字、标点、换行)本来就落在单字节 token 上,99.4\% 的汉字在原模型里从不走字节路径,所以零适应的模型能读却读不好:未训练的收缩检查点四域 bpb 为 zh 3.02 / en 4.43 / code 2.54。这组数只用来检验收缩没做错(与第一轮 27B 试点的零适应数同量级),不当作起点能力的度量。

\subsection{语料:对话类监督自产,知识阅读用人写的原文}
LD3-base 两段课程共读入 7.08 G 字节、受监督 4.77 G 字节。监督字节按来源分:wiki 原文 2.56 G(54\%)、词典 1.61 G(34\%)、字典 0.30 G(6\%)、一般对话 0.16 G、编码会话 0.14 G。要先说清楚:\textbf{占监督字节 54\% 的 wiki 是人类写的维基百科原文},按预训练方式首尾相接切窗读入,不是模型自产;自产的是其余四类对话 / 行为类监督。原文负责知识阅读,自产对话负责"模型用字节说自己的话"。

对话类监督全部由原版 Qwen3.8-27B 自己生成(on-policy),依据是两组对照。Ornith 试点线的六个"少思考"训练对照里,只有模型自己当前的输出训自己是零代价(自己的正确样本训 3.2 遍,正确率 49\% 不动),四个用别人写的数据的配方全部掉 5–14 分,与内容好坏无关;"当前"严格到差一个训练阶段就不算。跨词表蒸馏的三臂对照指向同一件事:监督别人写的提示词前缀让 code bpb 变差 26\%、en 变差 16\%,限制到本轮生成后损伤消失。所以本文 CE 只算助手轮(含思考);多轮会话\textbf{按轮独立补答},前缀原样保留,只监督本轮。

四类自产语料:(1)\textbf{字典},12,126 个汉字(8,105 个规范字 + 4,021 个简繁写法不同的繁体字),读音、义项、简繁关系取自 Unihan / CC-CEDICT / OpenCC,每字三轮对话(条目 $\rightarrow$ 整条译成另一语言 $\rightarrow$ 追问);它是"字 $\leftrightarrow$ 三个字节"最直接的显式监督,所以排第一。(2)\textbf{词典},141,361 个词条种子 $\times$ 4 轮。字典与词典的写法规则是"事实来自词典,文字由老师写":提示是自然对话,格式要求只放在不受监督的用户轮,人设摊开成 33 种以免文风被单一人设带偏。(3)\textbf{一般对话},六个公开会话池去重后 60,549 段、共 113 万个助手轮,其中 91\% 是带工具调用的 agent 轮。(4)\textbf{真人编码会话},两个公开数据集(6,772 段 + 1,064 段)按轮补答两遍共 88,743 轮。全部生成开思考、用官方采样参数(temperature 1.0 / top\_p 0.95 / top\_k 20)。

生成量远大于训练用量:一般对话本机生成了 30 多万轮,进入课程的监督字节只有 0.16 G,因为前缀长、只监督末轮。本机生成合计约 94.9 h $\approx$ 760 H200 GPU 时;走托管接口的部分(作文 2,048 篇、逻辑题约 11,800 道、编码会话)算力不可见,只报调用量。作文与逻辑题\textbf{没有进入 LD3-base 的两段课程}:逻辑题(reasoning-gym 99 类,10,560 道、答对 9,618 道)用在 3.4 节,作文(两批共约 2,560 篇)只生成、未入任何训练包。

\subsection{两段课程:先学编码,再读书}
训练一律全参、ZeRO-3,检查点只存权重(47 GB);两段都用常数学习率,段间仅权重续训(优化器重置、5 步预热),实测 bpb 无回退。决策同时看四域字节 bpb、30 题贪心探针和对话留出集,单看任一个都不做决定。

\textbf{M1 编码阶段}(64K 字节窗,2,077 步,lr 3e-6,47.5 h):字典 : 词典 : 一般对话按监督字节 1:1:1 混合,字典全量一遍,词典按字覆盖抽样,一般对话只收整条装得下 64K 的会话。实际读入 1.143 G 字节、监督 0.616 G 字节(训练记录里"各 0.28 GB、共 0.85 GB"是选样目标值,本文以包统计为准)。bpb 轨迹说明"中文才是新技能":en 100 步就到 0.58 并停住,zh 从 3.02 经 400 步 1.12 到收官 0.900。但 M1 收官时模型会读不会用:探针 29/30,术语"英 $\rightarrow$ 中"却 0/12,英文通路 11–12/12——知识在,坏的是中文词形与概念之间的连接。

\textbf{M2 读书阶段}(16K 字节窗,6,761 步,lr 5e-6,三段续训合计 131 h):wiki 原文 : 一般语料 : 词典 = 1:1:1 \textbf{按条数}混合,换成监督字节就是 wiki 54\% / 词典 34\% / 对话类 6\%(3.2 节),因为对话类前缀不算损失。实际读入 5.932 G、监督 4.156 G 字节。M2 之后术语"英 $\rightarrow$ 中"从 0/12 到 12/12(3,000 步时已达到),探针 30/30,GSM8K 对话 300 题 294;MMBench-CN 65.2\% $\rightarrow$ 72.2\%,纯文本训练没伤到冻结的视觉塔。

两段都\textbf{只混合、不先后}:试点线上"窄语料先训、正常数据后补"的配方,行为被推回来了,丢掉的能力却回不来;所以 M1 从第一步就混一般对话,M2 的 wiki 也和对话一起混。

\subsection{字节专用后训练:方法与负面经验}
LD3-base 没有做任何后训练。本节记录在它之上做的一条后训练链(记作 M3)的方法与结论;这条链的权重已删除、只剩日志,不进主表。

起因是部署后的观察:LD3-base 自产回答里 46\% 含 8-gram 复读(同题下另一个商用模型 6\%),而且写进系统提示的写作规范它不听。提示无效就只能训进去。M3 的目标按优先级是避免复读、简洁正确的思考、按规范工作,链条四步,全部 on-policy:

\begin{enumerate}
\item \textbf{规范 SFT}:规范一般对话 10.5K 条 + 规范逻辑题答对全收约 5K 条 + 回放 4K 条,16K 窗,lr 5e-6 WSD,259 步。
\item \textbf{自采,护栏关}:从训练包取"最后一轮回答之前"的前缀出题,每题采 4 条,官方采样参数,单条上限 8,192 字节。采样时关掉解码护栏让复读暴露,护栏掐断的样本进了训练数据就是教模型在答案段写 \texttt{\textless{}/think\textgreater{}}。
\item \textbf{挖样与 RFT}:干净 = 自己停、有 \texttt{\textless{}/think\textgreater{}}、8-gram 重复率 $\leq$ 0.03、无整句重复与磕巴、规范 lint 分 $\geq$ 80;差 = 顶上限、或重复率 $\geq$ 0.10、或整句重复 $\geq$ 2;逻辑题以"正确且思考最短"为 chosen。干净样本先做一轮 RFT(lr 3e-6)。
\item \textbf{长度归一化 DPO}:每题最多一对,chosen 取干净里 lint 最高,rejected 取差里重复率最高。奖励 r = $\beta$$\cdot$(log p $-$ log p\_ref)/len,损失 L = $-$log $\sigma$(r\_c $-$ r\_r $-$ $\gamma$) + $\alpha$$\cdot$($-$log p\_c/len\_c),$\beta$ 2.0 / $\gamma$ 0.5 / $\alpha$ 0.2,lr 5e-7。归一化是必要的:对子里 rejected 比 chosen 长的占 99\%(长度中位 634 对 4,918 字节)。
\end{enumerate}
三轮结果给出三条负面经验:

\begin{itemize}
\item \textbf{跳过 RFT 直接 DPO 伤精确推理。} 自采 61,600 条,差样本率 11.9\%;直接 DPO(4,460 对,140 步)把差样本率压到 7.7\%,但 GSM8K 300 题贪心对话从 97.7\% 掉到 80.7\%,输出均长 1,198 $\rightarrow$ 3,681 token。(97.7\% 是 300 题带思考贪心探针的口径,不是第 5 节主表的口径。)
\item \textbf{RFT 先于 DPO 能拉回来。} 补做"最短且对"RFT 再 DPO(7,004 对,219 步),GSM8K 300 题回到 92.7\%。
\item \textbf{继续压复读会把思考压没。} 第三轮验收差样本率只剩 0.2\%,但 252 条带思考回答里 211 条思考不到 60 个字符、中位 28 个字符,前两轮中位 136 与 210。0.2\% 大半来自"不思考"。根子在配对规则:chosen 取"最短"会系统性选到无思考样本;后来改成 chosen 思考长度 $\geq$ rejected 的 1/4、思考语言一致、总长度最接近,让一对之间除了"病"没有别的系统差异。 对数不够也不要训:218 对 28 步的 DPO pref\_acc 0.375、复读反升。
\end{itemize}
\subsection{深草稿 MTP}
字节模型每步只出一个字节,解码步数约为原版的 4 倍,多 token 预测(MTP)草稿头是补回速度的主要手段。Qwen3.5 系原生带一个 MTP 层:\texttt{fc[5120, 10240]} 拼接下一 token 嵌入与主干隐状态,接 1 层全注意力 decoder 层,与主干共享嵌入和 lm\_head(所以 384 词表直接适用),424.7M 参数;主干训完后要单独重建这一层。

草稿深度按汉字数定,不是 BPE 模型习惯的 2–3 步:一个汉字 3 字节,字内续字节近乎确定,默认 K=9(3 个汉字),之后开到 K=15(5 个汉字);训练代价 $\propto$ K,但 MTP 层只占主干的 1/64。LD3-base 的头冻主干只训头,K=15,1,500 步,lr 1e-4;实测后按"每轮接受 < 8 字节或提速 < 1.5$\times$"触发第二轮,用模型自己写的约 1 亿字节(60,000 题)on-policy 续训,lr 5e-5,留接受长度更高的头;两轮合计约 35 GPU 时。

两条坑。其一,训练口径与服务口径差很远:试点线 K=15 的头第 1 步命中 0.935,服务里每轮只接受 5.33 字节、只快 20\%,因为每位置准确率从 0.93 衰减到第 16 位的 0.40。LD3-base 的头在单卡以 K=8 运行时每轮接受 5.7 字节,16 并发 1.8$\times$、单流 3.3$\times$(5.6 节在 48G 单卡上未能复测的配置)。其二,主干一变旧头作废,必须重训;多层草稿模型与单步目标微调在我们这里都不及 1 层 MTP。

\subsection{解码护栏}
字节模型需要一层 BPE 模型不需要的解码护栏,原因在第 6 节展开。护栏实现为推理框架的自定义 logit 处理器,四个部件:

\begin{itemize}
\item \textbf{思考预算与答案预算}:\texttt{\textless{}think\textgreater{}} 之后到预算强制写 \texttt{\textless{}/think\textgreater{}};\texttt{\textless{}/think\textgreater{}} 之后到上限强制 \texttt{\textless{}\textbar{}im\_end\textbar{}\textgreater{}}——收口后答案段照样可能循环到顶,所以两段都要预算。服务默认各 16,384 字节。
\item \textbf{UTF-8 合法性掩码}:前导字节之后只允许续字节(128–191),字符边界禁止裸续字节与非法前导。
\item \textbf{循环熔断}:连续 1,024 步仍在重复才强制收口 / 结束,平时\textbf{不改 logits}。早期版本逐字节禁重复,试点线实测 41\% 的解码步闸门是开的(英文、LaTeX、代码里 16 字节内重复本来就常见),思考模式赚、非思考模式亏(GPQA 思考 48.0 $\rightarrow$ 60.6,非思考 60.6 $\rightarrow$ 53.0)。
\end{itemize}
两条使用规则。第一,\textbf{必须关掉推理框架的重叠调度}:重叠调度下护栏读到的输出滞后一个 token,模型刚吐出前导字节时护栏仍以为在字符边界,把合法续字节全禁掉,护栏自己制造了它要防的非法 UTF-8;三配置对照见 6.2 节。第二,护栏能救回的有上限:试点线三个护栏版本把"抽不到答案"从 77 题压到 41–53 题、收口率从 126 提到 198/198,答对数却全部是 120/198(BPE 原版 170);解码侧榨干之后剩下的是能力差距。所以护栏在"思考模式 + 有确定答案"的任务上是净收益,在要求长输出与特定结构的任务上会砍掉没写完的结构;自采训练数据时关护栏(3.4 节)。第 4 节评测里的护栏只阻止非法字节序列与超长输出,不改变合法输出的分布。

\section{实验设置}
\subsection{模型与检查点}
被比较的只有两个模型:

\begin{itemize}
\item \textbf{orig}:官方发布的 Qwen3.8-27B(FP8 权重,block-128 配方),字节级 BPE 词表 248,077 项(嵌入表 248,320 行)。
\item \textbf{LD3-base}:在同一主干上把词表收缩到 384(256 个字节符号 + 128 个特殊 / 保留 id),嵌入与输出头从各 1.27B 参数缩到约 2M(语言模型部分 26.9B $\rightarrow$ 24.4B),经两段课程(3.3 节)续训后的检查点,外加一个深度 8 的 MTP 草稿头(3.5 节)。本文所有表都用这一个检查点;为对齐口径,也量化成同配方的 block-128 FP8。
\end{itemize}
LD3-base 没有做过任何面向榜单的后训练;它是"字节化是否保住了原版能力"这个问题的最直接答案。字节模型特有的后训练问题(复读、贪心跑飞)在第 6 节单独讨论方法,不进主表。

\subsection{训练算力}
8$\times$H200,ZeRO-3 全参续训:M1(编码阶段)47.5 h,M2(读书阶段,三段合计)131 h,主干共 178.6 h $\approx$ 1,430 GPU 时;MTP 头约 35 GPU 时。实际读入 7.08 G 字节、受监督 4.77 G 字节(只对助手轮算损失)。语料生成另计:自产对话 / 字典 / 词典 / 逻辑题在本机八卡上约 760 GPU 时(按生成队列日志),连同外部 API 调用折算约 2,190 GPU 时(\texttt{E3\_E4.md})。按 6ND 粗估主干续训约 1e21 FLOPs,与 Bolmo 同量级;\textbf{数据量少一个数量级,算力不少},正文不写"便宜 N 倍"。

\subsection{评测口径(E1 / E2)}
两列在\textbf{同一张 RTX 6000 Ada 48G 上、同一套服务栈}(sglang v0.5.17,FP8,单卡,无草稿,非流式,\texttt{--disable-radix-cache})跑完全部榜单,避免把服务端差异算进模型差异。具体:

\begin{itemize}
\item \textbf{两档解码}:思考档 temperature 0.6 / top\_p 0.95,把 \texttt{\textless{}/think\textgreater{}} 之前的内容过滤掉再判分;非思考档 temperature 0。两档都通过 chat template 的 \texttt{enable\_thinking} 开关控制,模板与官方一致。
\item \textbf{输出上限按字节 ÷ 4 对齐}:字节模型按字节计上限,BPE 模型取其 1/4 的 token 数(思考档 32,768 B / 8,192 tok;非思考档 16,384 B / 4,096 tok;LongBench-v2 5,000 B / 1,250 tok)。实测中位输出长度字节 / token 比约 3(5.1 脚注),4 倍口径对字节模型不更紧。
\item \textbf{逐条计数伴随每个分数}:顶上限(\texttt{stop\_reason = max\_tokens})与思考档缺 \texttt{\textless{}/think\textgreater{}} 的条数一并给出,避免把"没写完"和"不会"混在一起。
\item \textbf{解码护栏}:字节模型服务端带 UTF-8 合法性掩码、思考 / 答案字节预算与熔断(3.6 节);BPE 模型不需要也没有。护栏只阻止非法字节序列与超长输出,不改变合法输出的分布。
\item \textbf{数据集}:GSM8K 测试全集(1,319)、MATH-500(500)、HumanEval+(164)、IFEval(541 题 / 834 条指令)、C-Eval val(1,346)、MMLU-Pro(14 科各抽 72 题共 1,008)、LongBench-v2(能放进上下文的 117 题)。判分用 evalscope 的默认判分器,两列同一份代码。
\item \textbf{bpb}:留出窗口集 4 域 $\times$ 96 窗 $\times$ 4,096 字节;字节模型直接对每字节 logprob 求和;BPE 模型对 token logprob 求和后除以同一段文本的字节数。
\item \textbf{单种子、单次运行}。每项给逐题自助法 95\% 区间与同题配对的 $\Delta$ 区间(2,000 次重采样,附录表 A2)。
\end{itemize}
\subsection{字符级任务口径(E5)}
\begin{itemize}
\item \textbf{题集}:CUTE(leukas/cute,14 项,每项抽 400 题)、EXECUTE(benjamin/execute)的 eng 与 zho 两语种(每项 200 题),以及自建中文字符题 15 项(每项 200 题;繁简互转各 107 题,受可一对一互转的自然句数量限制)。自建题的字表取自 12,126 字的常用字表,繁简事实取 OpenCC \texttt{STCharacters/TSCharacters}(只保留一对一的字),拼音 / 笔画 / 部首取 Unihan \texttt{kMandarin/kTotalStrokes/kRSUnicode},自然句取 C-Eval 题干(8–40 字、无 ASCII、$\geq$ 80\% 汉字,共 675 句)。剔除 extra\_* 后共 13,214 题。
\item \textbf{判分}:精确匹配。先去掉"答案:"前缀、首尾引号与句号,再按题型决定是否忽略空白;只取回答的第一个非空段落。思考档只看 \texttt{\textless{}/think\textgreater{}} 之后。
\item \textbf{解码}:非思考档 temperature 0 为主,全部题集两列都跑;思考档只补 CUTE 全 14 项与自建中文题各 100 题(temperature 0.6 / top\_p 0.95)。输出上限 1,024 字节 / 256 token(思考档 16,384 字节 / 4,096 token)。
\item \textbf{剔除}:EXECUTE-zho 的三个 \texttt{extra\_*} 拆分(\texttt{extra\_c2g}、\texttt{extra\_contains}、\texttt{extra\_g2c},部件图式 $\leftrightarrow$ 整字)在公开数据里只有"Task: ⿰亻吏"一类输入、没有任务说明(原意为少样本),两列零样本都得 0,\textbf{不进表}。EXECUTE-zho 因此与 eng 一样是 12 项。
\end{itemize}
\subsection{效率口径(E6)}
同一张卡、同一服务栈,四个配置:orig、orig + NEXTN 草稿、LD3-base、LD3-base + NEXTN 草稿(K = 8)。三个量:(1)单流解码,10 条中文与 10 条英文约 300 词的作文题,报\textbf{汉字 / 秒}与\textbf{英文字符 / 秒}而不是 token / 秒;(2)并发 16 的总吞吐,同样按字符计;(3)prefill:16,384 与 65,536 字节的中文窗口,max\_tokens = 1,报首字时间(TTFT),各 3 次取中位。草稿配置要把显存占比降到 0.80、上下文降到 65,536 才能在单卡起来;起不来的配置如实记"单卡起不来"。

\section{结果}
\subsection{主表:字节化后保住了多少}
表 1 是本文的主结果。两档合在一起看:

\begin{itemize}
\item \textbf{非思考档}(日常用法):7 项中 GSM8K($-$0.4)、MATH-500(+0.8)、HumanEval+($-$2.5)、MMLU-Pro($-$1.3)在 $\pm$2.5 以内;C-Eval $-$4.7;IFEval 四个指标 $-$4.1 \textasciitilde{} $-$6.2 是最大的损失。
\item \textbf{思考档}:MATH-500(+1.4)与 IFEval 四项(+0.2 \textasciitilde{} +0.6)持平或略高,HumanEval+ $-$1.8、MMLU-Pro $-$1.3;GSM8K $-$4.0 与 C-Eval $-$4.5 是两处明显的损失。
\item \textbf{最稳定的损失是中文考试}(C-Eval 两档都约 $-$4.5)\textbf{与非思考档的指令遵循};\textbf{最稳定的保留是数学与 MMLU-Pro}。LD3-base 没做过任何指令类后训练,非思考档 IFEval 的差距有一部分应归于此(思考档同一项持平),但本文没有分离这个因素的实验。
\end{itemize}
\begin{table}[htbp]
\centering
\footnotesize
\caption{主表:orig 与 LD3-base 在同一张 RTX 6000 Ada 48G、同一服务栈下的分数(单种子;顶上限与缺 \texttt{\textless{}/think\textgreater{}} 条数见表注;配对自助法区间见附录表 A2)}
\label{tab:main}
\begin{adjustbox}{max width=\textwidth}
\setlength{\tabcolsep}{3.5pt}
\begin{tabular}{llrrrrrr}
\toprule
数据集(n) & 指标 & orig 思考 & \textbf{LD3 思考} & $\Delta$ & orig 非思考 & \textbf{LD3 非思考} & $\Delta$ \\
\midrule
GSM8K(1319) & acc & 97.0 & 93.0 & $-$4.0 & 96.4 & 96.0 & $-$0.4 \\
MATH-500(500) & acc & 94.0 & 95.4 & +1.4 & 92.8 & 93.6 & +0.8 \\
HumanEval+(164) & pass@1 & 90.2 & 88.4 & $-$1.8 & 92.7 & 90.2 & $-$2.5 \\
IFEval(541) & prompt-strict & 87.8 & 88.2 & +0.4 & 81.9 & 76.2 & $-$5.7 \\
IFEval(541) & prompt-loose & 89.6 & 90.2 & +0.6 & 86.0 & 79.8 & $-$6.2 \\
IFEval(834) & inst-strict & 91.6 & 91.8 & +0.2 & 87.4 & 83.3 & $-$4.1 \\
IFEval(834) & inst-loose & 92.9 & 93.4 & +0.5 & 90.8 & 86.2 & $-$4.6 \\
C-Eval val(1346) & acc & 89.9 & 85.4 & $-$4.5 & 88.1 & 83.4 & $-$4.7 \\
MMLU-Pro(1008) & acc & 80.1 & 78.8 & $-$1.3 & 79.4 & 78.1 & $-$1.3 \\
LongBench-v2(117)\textdagger{} & acc & 27.4 & 39.3 & +11.9 & 42.7 & 53.8 & +11.1 \\
\bottomrule
\end{tabular}
\end{adjustbox}
\end{table}
表 1 的表注:

\begin{itemize}
\item \textdagger{} LongBench-v2 两列两档都被输出上限主导(顶上限:思考档 orig 86/117、LD3-base 69/117;非思考档 48/117、34/117),且 1,250 token 对 BPE 模型比 5,000 字节对字节模型更紧(中文 1 token $\approx$ 1.5–2 字节),\textbf{这一行不作为字节化优势的证据},只说明同上限下字节模型更早收口。
\item 顶上限 / 缺 \texttt{\textless{}/think\textgreater{}}(思考档,orig / LD3-base):GSM8K 9/4、MATH-500 28/15、HumanEval+ 13/6、IFEval 28/27、C-Eval 48/31、MMLU-Pro 138/93;非思考档顶上限:GSM8K 7/1、MATH-500 27/9、HumanEval+ 0/0、IFEval 5/22、C-Eval 30/37、MMLU-Pro 61/29。字节模型在每一项思考档的顶上限条数都不多于原版,没有"思考不停";原版思考档 MMLU-Pro 13.7\% 顶满,其 80.1 可能被 8,192 token 上限略压低。
\item 中位输出长度(思考档)orig token / LD3-base 字节:GSM8K 289 / 870、MATH-500 607 / 1,382、MMLU-Pro 712 / 2,263,比约 3。
\item 单种子;逐题自助法 95\% 区间(2,000 次,配对 $\Delta$):配对区间不含 0 的只有 GSM8K 思考档 [$-$5.3, $-$2.9]、C-Eval 思考 [$-$6.3, $-$2.8] / 非思考 [$-$6.5, $-$3.0] 与上限主导的 LongBench-v2;MATH-500、HumanEval+、MMLU-Pro、GSM8K 非思考的 $\Delta$ 区间都含 0(全表见附录表 A2)。重采样只覆盖题目抽样方差,不覆盖解码随机性。
\item 用时(同卡):orig 两档 7 h 44 min,LD3-base 两档 35 h 57 min($\approx$ 4.6$\times$),见 5.6。
\end{itemize}
\subsection{语言建模:bpb}
表 2。四域都低约 2 个标准误,但窗口集是 LD3 训练时的留出集,与其训练数据同分布而原版没见过,所以\textbf{不能解读为建模能力更强},只能说字节化后在自家分布上回到并略低于原版水平;zh\_ood 一行最可信。图 1 是训练全程的曲线:零适应起点(词表收缩后未训练)zh 3.02 / zh\_ood 3.29 / en 4.43 / code 2.54;M1 100 步时 en 已到 0.58 而 zh 还在 2.02,M1 收官 zh 0.900 / zh\_ood 1.213 / en 0.575 / code 0.203;M2 阶段四域单调缓降,收官 0.740 / 0.995 / 0.547 / 0.191。训练日志是 bf16 权重、训练器进程内算 CE,E2 是 FP8 权重经服务取 logprob,同一检查点两种读法相差 $\leq$ 0.002,所以表 2 与图 1 可以对读。英文在 M1 的前 100 步就回到原版水平,中文用了全部 7 G 字节才追到——这就是 3.3 节"中文才是新技能"在损失上的样子。

\begin{figure}[htbp]
\centering
\begin{tikzpicture}
\begin{axis}[
  width=0.92\textwidth, height=6.2cm,
  xlabel={累计实读字节(G,去填充;M1 与 M2 段内按步线性折算)},
  ylabel={bpb},
  ymode=log, ymin=0.15, ymax=5,
  log ticks with fixed point,
  ytick={0.2,0.3,0.5,0.75,1,1.5,2,3,5},
  xmin=-0.1, xmax=7.3,
  grid=major, grid style={gray!25},
  legend style={at={(0.98,0.97)},anchor=north east,font=\footnotesize,cells={anchor=west}},
  tick label style={font=\footnotesize}, label style={font=\small},
  every axis plot/.append style={thick},
]
\addplot[blue!70!black]  table[x=bytes,y=zh]     {figs/bpb_m1.dat}; \addlegendentry{zh}
\addplot[blue!70!black]  table[x=bytes,y=zh]     {figs/bpb_m2.dat};
\addplot[red!70!black]   table[x=bytes,y=zh_ood] {figs/bpb_m1.dat}; \addlegendentry{zh\_ood}
\addplot[red!70!black]   table[x=bytes,y=zh_ood] {figs/bpb_m2.dat};
\addplot[green!50!black] table[x=bytes,y=en]     {figs/bpb_m1.dat}; \addlegendentry{en}
\addplot[green!50!black] table[x=bytes,y=en]     {figs/bpb_m2.dat};
\addplot[orange!90!black] table[x=bytes,y=code]  {figs/bpb_m1.dat}; \addlegendentry{code}
\addplot[orange!90!black] table[x=bytes,y=code]  {figs/bpb_m2.dat};
\addplot[black,dashed,thin] coordinates {(-1,-1) (-1,-1)}; \addlegendentry{orig(BPE 读法)}
\addplot[blue!70!black,dashed,thin]   coordinates {(-0.1,0.778) (7.3,0.778)};
\addplot[red!70!black,dashed,thin]    coordinates {(-0.1,1.022) (7.3,1.022)};
\addplot[green!50!black,dashed,thin]  coordinates {(-0.1,0.587) (7.3,0.587)};
\addplot[orange!90!black,dashed,thin] coordinates {(-0.1,0.205) (7.3,0.205)};
\addplot[black,dashed,thin,forget plot] coordinates {(1.143,0.15) (1.143,5)};
\node[font=\footnotesize,anchor=south west] at (axis cs:1.2,3.2) {M1 $\rightarrow$ M2};
\end{axis}
\end{tikzpicture}
\caption{四域留出窗口 bpb 随训练字节数的变化(实线:LD3,训练日志 bf16 读数;虚线:orig 按 token logprob 折算到字节)。英文与代码在 M1 的前 100 步就回到原版水平,中文到 M2 收官才追到并略低于原版;竖虚线为 M1 / M2 分界(1.143 G)。终点与表 2 的 FP8 服务读数相差 $\leq 0.002$。}
\label{fig:bpb}
\end{figure}
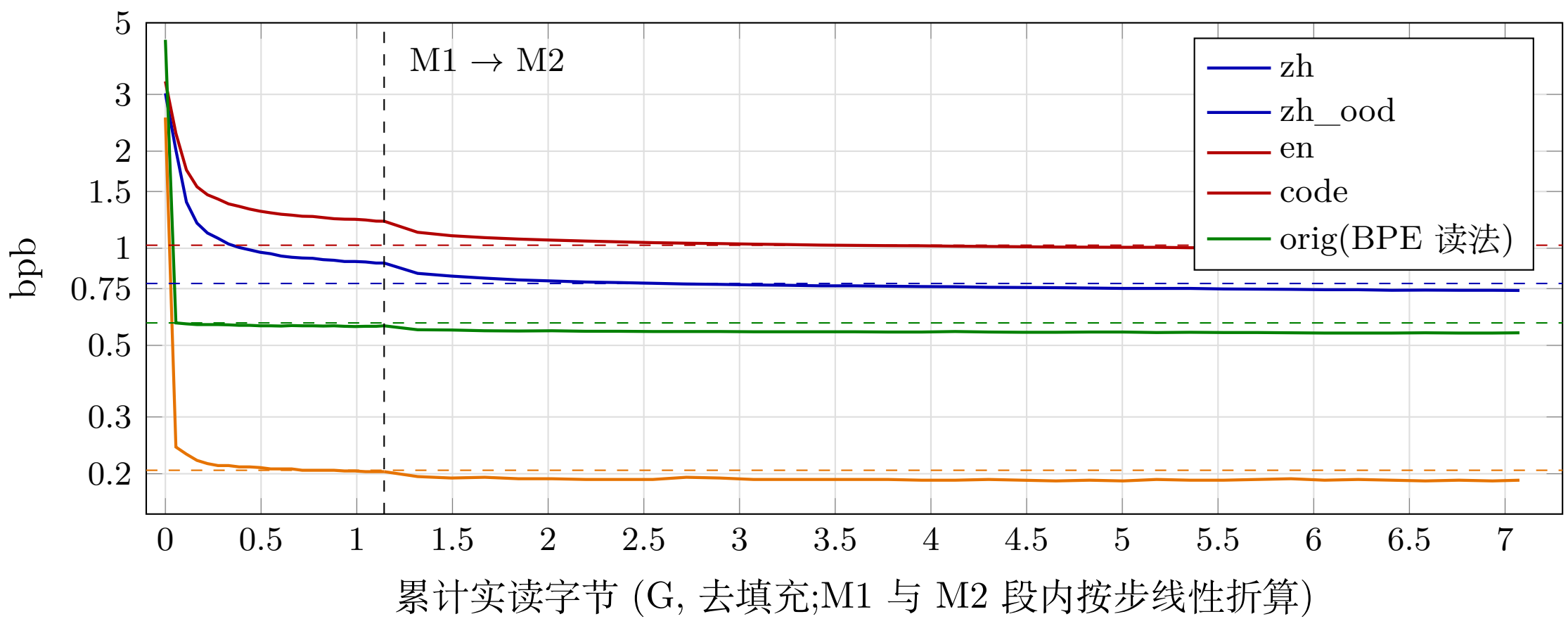

\begin{table}[htbp]
\centering
\small
\caption{留出窗口集 bpb(4 域 $\times$ 96 窗 $\times$ 4,096 字节;$\pm$ 为窗间标准误;orig 按 token logprob 之和除以字节数)}
\label{tab:bpb}
\begin{tabular}{lrrr}
\toprule
域 & orig(BPE 读法) & \textbf{LD3-base} & $\Delta$ \\
\midrule
zh(中文网文) & 0.778 $\pm$ 0.021 & 0.742 $\pm$ 0.018 & $-$0.036 \\
zh\_ood(中文,域外) & 1.022 $\pm$ 0.013 & 0.997 $\pm$ 0.013 & $-$0.025 \\
en & 0.587 $\pm$ 0.017 & 0.549 $\pm$ 0.014 & $-$0.038 \\
code & 0.205 $\pm$ 0.008 & 0.190 $\pm$ 0.008 & $-$0.015 \\
\bottomrule
\end{tabular}
\end{table}
\subsection{中文}
字节化最可能伤的是中文,所以 M1 / M2 之间用两组小探针盯着它,并在终点补一项全字复述。

\begin{itemize}
\item \textbf{术语五问}。12 个中英术语(梯度下降、反向传播、哈希表、死锁等)各 5 种问法:A 英 $\rightarrow$ 中只答术语、B 中 $\rightarrow$ 英、C 中问中答、D 英问要求中文答、E 中问括号给英文;贪心、带思考、上限 8,192 字节,人工判读答案段。M1 收官 A 0/12、B 6/12、C 6/12、D 11/12、E 12/12:知识在、英文通路通,坏的是"中文词形 $\leftrightarrow$ 概念"的连接,这是 3.3 节"中文才是新技能"的直接证据。M2 读到 3,000 步 A 即 12/12、B 10/12,之后不变。
\item \textbf{30 题贪心探针}。英 10 / 中 5 / 代码 5 / 格式 6 / 梯度 4,temperature 0、上限 16,384 字节,只判 \texttt{\textless{}/think\textgreater{}} 之后的答案段,代码题真跑测试用例。M1 29/30 $\rightarrow$ M2@3000 28 $\rightarrow$ M2@6000 30 $\rightarrow$ LD3-base 30/30。两组探针只用于训练中决策,n 太小,不作为主结果。
\item \textbf{12,126 字全字复述}(E11)。字表与训练用字典同源(规范字 8,105 + 繁体 4,021);每 20 字一组共 607 组,提示"原样复述、一字不漏",贪心、不思考,输出与输入逐字对齐,记漏字 / 错字 / 非法字节;两列同卡同口径,各跑字典顺序与打乱重分组两遍。两遍合计 24,252 字次,orig 错 3 字(汨 $\rightarrow$ 汩,以及两个生僻字抄成形近字),LD3-base 错 1 字(欽 抄成同组前面的形近字),全部是抄成邻近或形近字,没有漏字;两列都零非法 UTF-8、零顶上限。含 4,021 个繁体与扩展区生僻字在内,字节模型逐字节写字没有比原版差,差别在噪声内。
\end{itemize}
\subsection{字符级任务:字节的收益(E5)}
要回答的问题:BPE 模型看不见拼写,字节模型天生看得见——这个理论上的优势在 27B 上、在中文上有多大?表 3 给四个题集的宏平均(非思考档全部题目,思考档只补 CUTE 与自建题各 100 题/项),表 4 给单项。

\begin{table}[htbp]
\centering
\footnotesize
\caption{字符级任务宏平均(精确匹配 \%;非思考档全部题目,思考档每项 100 题)}
\label{tab:e5macro}
\begin{adjustbox}{max width=\textwidth}
\setlength{\tabcolsep}{3.5pt}
\begin{tabular}{llrlrrl}
\toprule
题集 & 项数 & orig 非思考 & \textbf{LD3-base 非思考} & $\Delta$ & orig 思考 & \textbf{LD3-base 思考} \\
\midrule
CUTE(英) & 14 & 75.2 & \textbf{84.0} & +8.8 & 89.4 & \textbf{91.2} \\
EXECUTE-eng & 12 & 76.0 & \textbf{83.2} & +7.2 & — & — \\
EXECUTE-zho & 12(剔 extra\_*) & 81.6 & \textbf{93.9} & +12.2 & — & — \\
自建中文字符题 & 15 & 54.4 & \textbf{64.0} & +9.7 & 87.8 & \textbf{89.1} \\
\bottomrule
\end{tabular}
\end{adjustbox}
\end{table}
非思考档(temperature 0,上限 1,024 字节 / 256 token,两列错误 0)的读法:

\begin{itemize}
\item \textbf{字符操作全部领先,中文幅度大于英文}。删 / 插 / 替换 / 交换字符、拼写变体(orth)、找字这几类,LD3-base 领先 +10 \textasciitilde{} +43:CUTE sub\_char 37.2 $\rightarrow$ 71.2、ins\_char 21.2 $\rightarrow$ 44.2;EXECUTE-zho swap\_char 48.5 $\rightarrow$ 91.5、ins\_char 65.5 $\rightarrow$ 99.5;自建题 zh\_swap\_char 13.0 $\rightarrow$ 50.5、zh\_del\_char 62.0 $\rightarrow$ 94.5、zh\_sub\_char 53.0 $\rightarrow$ 85.5。词级操作(del\_word / ins\_word / sub\_word / contains\_word)两列本来都接近满分,持平。
\item \textbf{不是靠多写}。字节模型的输出字节数约为原版 token 数的 2–4 倍,与序列长度比一致;顶上限条数两列都接近 0。原版要到思考档才能把这些题做对(CUTE 75.2 $\rightarrow$ 89.4、自建 54.4 $\rightarrow$ 87.8),代价是每题几百到几千 token(CUTE del\_word 平均 3,306 token、36/100 顶上限)。
\item \textbf{汉字知识项字节模型掉}。zh\_homophone(四选一选同音字)96.5 $\rightarrow$ 57.5、zh\_strokes 63.0 $\rightarrow$ 47.0、zh\_pinyin 98.0 $\rightarrow$ 91.5;zh\_radical +5.0、繁简互转 +1.9 持平。注拼音单字尚可,但四字比对读音明显不行,错答 85 个里 71 个选 A(拿不准就选第一项)。读音 / 笔画是"字 $\rightarrow$ 属性"的查表知识,字节化后每个汉字的 3 个字节要重新学,这类知识没有完全回来,与 5.1 中 C-Eval $-$4.5 同一方向。这是本节最该诚实写的一条:\textbf{字节化换来的是操作能力,付出的是一部分字级知识}。
\item \textbf{计数与定位两列都弱}。zh\_count / zh\_nth / zh\_index / zh\_reverse 原版 6–23、字节模型 25–37,非思考贪心下两个 27B 模型都数不准位置,字节化只是少错一点。
\item 剔除项:EXECUTE-zho 的 extra\_c2g / extra\_contains / extra\_g2c 没有任务说明(4.4 节),orig 63.5 / 0 / 0,LD3-base 15.5 / 0 / 0,不进表。
\end{itemize}
表 4 单项(非思考档,n:CUTE 400、EXECUTE 200、自建 200,繁简 107):

\begin{table}[htbp]
\centering
\footnotesize
\caption{字符级任务单项(非思考档,精确匹配 \%;n:CUTE 400、EXECUTE 200、自建 200,繁简互转 107)}
\label{tab:e5task}
\begin{adjustbox}{max width=\textwidth}
\setlength{\tabcolsep}{3.5pt}
\begin{tabular}{llrrrllrrr}
\toprule
题集 & 项 & orig & \textbf{LD3-base} & $\Delta$ & 题集 & 项 & orig & \textbf{LD3-base} & $\Delta$ \\
\midrule
CUTE & contains\_char & 83.8 & 94.0 & +10.2 & EXECUTE-zho & contains\_char & 99.5 & 100.0 & +0.5 \\
CUTE & contains\_word & 100.0 & 99.8 & $-$0.2 & EXECUTE-zho & contains\_word & 99.5 & 100.0 & +0.5 \\
CUTE & del\_char & 67.5 & 82.8 & +15.2 & EXECUTE-zho & del\_char & 96.0 & 99.0 & +3.0 \\
CUTE & del\_word & 96.8 & 96.2 & $-$0.5 & EXECUTE-zho & del\_word & 91.0 & 94.5 & +3.5 \\
CUTE & ins\_char & 21.2 & 44.2 & +23.0 & EXECUTE-zho & ins\_char & 65.5 & 99.5 & +34.0 \\
CUTE & ins\_word & 98.0 & 98.2 & +0.2 & EXECUTE-zho & ins\_word & 56.0 & 79.0 & +23.0 \\
CUTE & orth & 68.0 & 84.5 & +16.5 & EXECUTE-zho & spell & 100.0 & 100.0 & 0.0 \\
CUTE & sem & 95.8 & 94.2 & $-$1.5 & EXECUTE-zho & spell\_inverse & 100.0 & 100.0 & 0.0 \\
CUTE & spell & 100.0 & 100.0 & 0.0 & EXECUTE-zho & sub\_char & 83.5 & 93.0 & +9.5 \\
CUTE & spell\_inverse & 100.0 & 100.0 & 0.0 & EXECUTE-zho & sub\_word & 91.5 & 98.0 & +6.5 \\
CUTE & sub\_char & 37.2 & 71.2 & +34.0 & EXECUTE-zho & swap\_char & 48.5 & 91.5 & +43.0 \\
CUTE & sub\_word & 96.5 & 95.5 & $-$1.0 & EXECUTE-zho & swap\_word & 48.5 & 72.0 & +23.5 \\
CUTE & swap\_char & 11.0 & 22.5 & +11.5 & 自建 & zh\_count & 12.0 & 36.5 & +24.5 \\
CUTE & swap\_word & 76.8 & 92.8 & +16.0 & 自建 & zh\_reverse & 6.0 & 25.5 & +19.5 \\
EXECUTE-eng & contains\_char & 94.0 & 96.0 & +2.0 & 自建 & zh\_nth & 23.0 & 24.5 & +1.5 \\
EXECUTE-eng & contains\_word & 100.0 & 100.0 & 0.0 & 自建 & zh\_index & 22.0 & 35.0 & +13.0 \\
EXECUTE-eng & del\_char & 73.5 & 88.5 & +15.0 & 自建 & zh\_contains & 81.0 & 90.5 & +9.5 \\
EXECUTE-eng & del\_word & 97.5 & 96.0 & $-$1.5 & 自建 & zh\_del\_char & 62.0 & 94.5 & +32.5 \\
EXECUTE-eng & ins\_char & 11.5 & 40.5 & +29.0 & 自建 & zh\_ins\_char & 40.5 & 68.0 & +27.5 \\
EXECUTE-eng & ins\_word & 97.5 & 98.5 & +1.0 & 自建 & zh\_sub\_char & 53.0 & 85.5 & +32.5 \\
EXECUTE-eng & spell & 99.5 & 100.0 & +0.5 & 自建 & zh\_swap\_char & 13.0 & 50.5 & +37.5 \\
EXECUTE-eng & spell\_inverse & 99.5 & 99.5 & 0.0 & 自建 & zh\_s2t & 77.6 & 79.4 & +1.9 \\
EXECUTE-eng & sub\_char & 42.5 & 67.0 & +24.5 & 自建 & zh\_t2s & 96.3 & 98.1 & +1.9 \\
EXECUTE-eng & sub\_word & 99.5 & 99.0 & $-$0.5 & 自建 & zh\_pinyin & 98.0 & 91.5 & $-$6.5 \\
EXECUTE-eng & swap\_char & 11.5 & 25.0 & +13.5 & 自建 & zh\_homophone & 96.5 & 57.5 & $-$39.0 \\
EXECUTE-eng & swap\_word & 85.5 & 88.5 & +3.0 & 自建 & zh\_strokes & 63.0 & 47.0 & $-$16.0 \\
 &  &  &  &  & 自建 & zh\_radical & 71.5 & 76.5 & +5.0 \\
\bottomrule
\end{tabular}
\end{adjustbox}
\end{table}
思考档(每项 100 题,temperature 0.6 / top\_p 0.95,上限 16,384 字节 / 4,096 token)两列都接近满分,差距缩小到 CUTE +1.9、自建 +1.3,但分项仍有结构:

\begin{itemize}
\item \textbf{字符操作字节模型仍高} +4 \textasciitilde{} +34:zh\_reverse 59 $\rightarrow$ 93、swap\_word 77 $\rightarrow$ 96、zh\_swap\_char 84 $\rightarrow$ 98、ins\_char 85 $\rightarrow$ 95;中文计数 / 定位四项字节模型 100 / 100 / 100 / 93(原版 93 / 96 / 95 / 59)。
\item \textbf{汉字知识项思考补不回来}:笔画 34 vs 69、拼音 88 vs 99、同音 88 vs 99、部首 64 vs 72——查表知识不是推理能补的。
\item \textbf{两项字节模型反而掉}:CUTE del\_word 64 $\rightarrow$ 56、orth 67 $\rightarrow$ 48,原因是顶上限(43/100、53/100,原版 36 / 33),平均输出约 1 万字节。补跑重采后顶上限条数还涨了,说明是概率性的"思考绕不出来",不是个别题。这是第 6 节"字节模型特有问题"的直接证据,也是思考档结论不如非思考档干净的原因。
\end{itemize}
思考档单项见附录表 A1。图 2 把表 4 的 $\Delta$ 按项画出,操作类与知识类分色。

\begin{figure}[htbp]
\centering
\begin{tikzpicture}
\begin{axis}[
  width=0.95\textwidth, height=7.2cm,
  ybar, bar width=4.2pt,
  ylabel={$\Delta$ 精确匹配(LD3-base $-$ orig,百分点)},
  ymin=-45, ymax=48,
  xtick=data, xticklabels from table={figs/delta.dat}{label},
  x tick label style={rotate=90,anchor=east,font=\scriptsize,yshift=-2pt},
  ytick={-40,-20,0,20,40},
  grid=major, grid style={gray!25},
  legend style={at={(0.02,0.97)},anchor=north west,font=\footnotesize,cells={anchor=west}},
  tick label style={font=\footnotesize}, label style={font=\small},
  enlarge x limits=0.015,
  axis on top,
]
\addplot[fill=blue!60!black,draw=none]   table[x=idx,y=op]   {figs/delta.dat}; \addlegendentry{字符 / 词操作}
\addplot[fill=red!70!black,draw=none]    table[x=idx,y=know] {figs/delta.dat}; \addlegendentry{汉字知识(读音 / 笔画 / 部首 / 繁简)}
\addplot[fill=gray!60,draw=none]         table[x=idx,y=cnt]  {figs/delta.dat}; \addlegendentry{计数 / 定位}
\addplot[black,thin,forget plot] coordinates {(-1,0) (60,0)};
\end{axis}
\end{tikzpicture}
\caption{字符级任务单项 $\Delta$(非思考档,表 4 的数据)。左起依次为 CUTE(14 项)、EXECUTE-eng(12)、EXECUTE-zho(12)、自建中文题(15)。操作类几乎全部为正、中文幅度大于英文;汉字知识项(同音、笔画、拼音)为负;计数 / 定位两列都弱、字节模型少错一点。}
\label{fig:delta}
\end{figure}
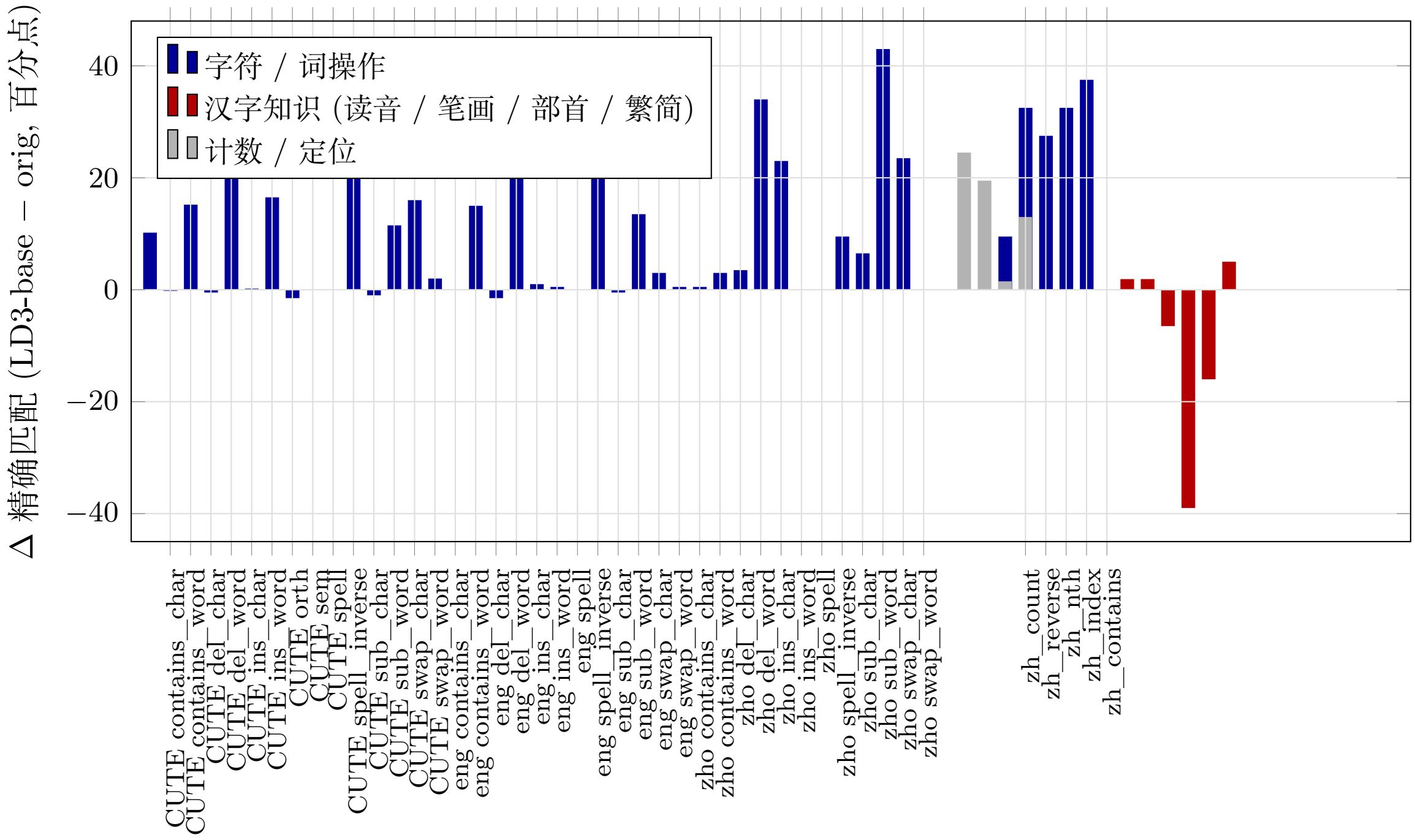

\subsection{长上下文}
LD3-base 的训练窗只有 64K(M1)与 16K(M2)字节,位置编码未动,也没有专门训长位置;本节问"原版的长上下文能力在字节化后剩多少"。四项都在 LD3-base 上测,\textbf{没有原版对照}(原版在同一题集上只留下"到 1M 不掉分"的记录,没有分数),所以只说明保留、不作比较。

\begin{itemize}
\item \textbf{128k 留出 bpb 0.257}(0–8k 桶 0.196,64–128k 桶 0.268):95 场留出长会话,128k 字节整窗一次前向,只算助手段逐位置 CE,按目标位置分桶。这个窗集与 5.2 的四域窗不是一套,不可横比。
\item \textbf{针测 8k–128k 全 4/4}:自有文档拼成干草堆,在深度 0.1 / 0.5 / 0.9 插一句"项目编号 $\rightarrow$ 口令"备忘,末尾提问,贪心,命中 = 口令出现。
\item \textbf{顺序探针 16k–128k 全 6/6}:相距 0.3 / 0.6 的两条备忘,问"哪条先出现"。
\item \textbf{到 1M 字节}:自建 8 类题(针、逐字抄录、hex / base64 / URL 编解码、两跳、求和、顺序、长文阅读),每档 46 题,证据在开头、问题在末尾、档间只差填充长度;贪心带思考、上限 16,384 字节,只判答案段。LD3-base 开静态 YaRN $\times$4.5 后 256K 39、512K 35、768K 37、1M 33/46;不开 YaRN 1M 27/46;错题主体是编解码类。
\end{itemize}
结论只有一句:在超出训练窗 2–64 倍的位置上模型仍能读、能定位,原版的位置能力经字节化保留下来,长度外推沿用原版的 YaRN 即可。E1 的 LongBench-v2 一行被输出上限主导,不作为长上下文证据(5.1 脚注)。

\subsection{效率:字节的代价(E6)}
序列长约 3–4 倍,代价不在"每步算得慢",而在"每步产出少"。表 5 是同一张 RTX 6000 Ada 48G、同一服务栈下的实测(单流 10 条 / 语种取均值,并发 16,prefill 3 次取中位;4.5 节口径)。

\begin{table}[htbp]
\centering
\footnotesize
\caption{同卡效率(RTX 6000 Ada 48G,sglang,FP8;单流为 10 条 / 语种均值,并发 16,prefill 为 3 次中位首字时间)}
\label{tab:eff}
\begin{adjustbox}{max width=\textwidth}
\setlength{\tabcolsep}{3.5pt}
\begin{tabular}{llllllll}
\toprule
配置 & 单流 tok/s & 单流 汉字/s & 单流 英文字符/s & 并发16 汉字/s & 并发16 英文字符/s & TTFT 16K B & TTFT 64K B \\
\midrule
orig & 22.9 & 33.1 & 115.4 & 482 & 1,441 & 1.16 s & 4.52 s \\
\textbf{LD3-base} & 24.3 & \textbf{7.7} & \textbf{24.3} & \textbf{97} & \textbf{332} & \textbf{5.83 s} & \textbf{29.57 s} \\
比值(LD3-base / orig) & 1.06 & 0.23 & 0.21 & 0.20 & 0.23 & 5.0$\times$ & 6.5$\times$ \\
orig + NEXTN 草稿 & 单卡起不来 &  &  &  &  &  &  \\
LD3-base + NEXTN 草稿 K=8 & 单卡起不来 &  &  &  &  &  &  \\
\bottomrule
\end{tabular}
\end{adjustbox}
\end{table}
读法:

\begin{itemize}
\item \textbf{每步速度相同}。单流 token/s 两列几乎一样(22.9 vs 24.3),这是同一主干的直接体现;并发 16 的总 token/s 也相近(329 vs 307)。差别全在每步产出:原版 1 token $\approx$ 1.45 个汉字或 5.0 个英文字符,字节模型 1 步 = 1 字节 $\approx$ 0.32 个汉字或 1 个英文字符。于是汉字 / 秒与英文字符 / 秒都降到原版的 1/4–1/5,并发下比例不变。
\item \textbf{prefill 更亏}。同样 65,536 字节的中文输入,原版约 2 万 token,字节模型 65,622 token,首字时间 4.52 $\rightarrow$ 29.57 s(6.5$\times$),比 token 数之比(约 3.2)还大一倍。多出来的部分是工程差距:字节模型走的是外挂的 GDN 分块与注意力核,没有 BPE 路径那样的调优。这不是模型本身的性质,但它是现状。
\item \textbf{草稿头在 48G 单卡上装不下}。两个 NEXTN 配置都起不来(27B FP8 权重 + 草稿头后 KV 池放不下),所以表 5 没有草稿行。MTP 对字节模型格外有效的依据只能引另一硬件上的记录:K = 8 每轮接受约 5.7 字节,单流约 3.3$\times$、16 并发约 1.8$\times$。按此折算,单流汉字 / 秒可回到原版的 0.7–0.8 倍,并发下回到 0.4 倍左右——\textbf{即使加草稿,同硬件吞吐仍低于原版}。
\item 全程口径的另一个数:E1 两档全部榜单同卡用时 orig 7 h 44 min、LD3-base 35 h 57 min(4.6$\times$),与表 5 的单项比值一致。
\end{itemize}
\subsection{多语种(E11)}
LD3-base 只用中英语料续训(3.2 节)。本节用 22 种语言 $\times$ 3 项的冒烟测试看"字节通道"与"生成能力"分别剩多少,两列同卡同口径(E11):(1)原样复述一句该语言(NFC 归一后逐字比对);(2)把同一句中文译成该语言;(3)用该语言问一道首都类常识题、要求该语言作答;贪心,不思考(上限 512 / 512 / 1,024)与思考(4,096 / 4,096 / 8,192)各一遍,两列 max\_tokens 数值相同(对 BPE 模型相当于约 3–4 倍文本预算,原版的数字因此是上界)。翻译与问答自动判分:输出字母落在目标文字区段的比例(script\_match)、8-gram 重复率与顶上限判循环、问答看是否含金标;保住 = 译与答都 script\_match $\geq$ 0.9 且不循环,退化 = 任一项循环或 script\_match < 0.5,其余为勉强。

\begin{table}[htbp]
\centering
\small
\caption{22 语种冒烟(E11;复述为逐字正确的语言数,问答为答对数,保住 / 勉强 / 退化为按文字系统与循环自动分档的语言数)}
\label{tab:ml}
\begin{tabular}{lrrrrr}
\toprule
模型 / 档 & 复述 & 问答答对 & 保住 & 勉强 & 退化 \\
\midrule
orig 非思考 & 22/22 & 22/22 & 21 & 1 & 0 \\
orig 思考 & 21/22 & 22/22 & 22 & 0 & 0 \\
\textbf{LD3-base} 非思考 & 22/22 & 13/22 & 11 & 0 & 11 \\
\textbf{LD3-base} 思考 & 22/22 & 14/22 & 13 & 0 & 9 \\
\bottomrule
\end{tabular}
\end{table}
读法:

\begin{itemize}
\item \textbf{字节通道完好,生成能力按文字系统分化。} 复述两列都接近全对(原版思考档泰语一句声调符号重复,是 BPE 在泰文上的 token 边界毛病;字节模型 22/22)。生成上原版 22 种全保住;字节模型只保住拉丁与西里尔文字的大语种加日语(非思考档 11 种:俄、日、阿拉伯、德、法、西、葡、越、土、波兰、意),希腊、韩、波斯、印度系、泰、希伯来等非拉丁文字退化:译文循环或词语乱堆,问答答非所问(印地语说印度的首都是"印地语")。
\item \textbf{退化是语料问题,不是架构问题。} 复制完好说明字节通道没坏;缺的是这些语言的监督字节,两段课程里只有中英文(3.2 节)。要把它们补回来需要在续训语料里掺对应语言,本文没做。
\item \textbf{思考对小语种是双刃。} 希腊、韩、乌克兰、印尼 4 种开思考后从退化回到保住(译文一次写完);越南、斯瓦希里、阿拉伯 3 种反而掉下去,原因全是思考本身在该语言里循环到上限、正文为空;净效果 +2 种(11 $\rightarrow$ 13),用时 1,356 s 对原版 171 s。这与 5.4 / 6.1 的"思考绕不出来"是同一件事。
\end{itemize}
\section{字节模型特有问题}
把一个 BPE token 摊成约 4 个字节,等于把每一次采样决定拆成 4 次。下面四个问题都从这里来;它们在 BPE 模型上不存在或不严重,在字节模型上必须单独处理。

\subsection{贪心解码下跑飞}
\textbf{现象。} temperature 0 下字节模型会写到输出上限还不收口。试点线一次非思考 GPQA-Diamond 贪心评测:198 题里 34 题(17\%)写满 16,384 字节上限、没有答案行;同一套权重的 BPE 原版贪心 0 条抽不到答案、输出 p90 只有 2,005 字符。第一轮 27B 字节化的思考版贪心评测里,IFEval 42\% 的样本顶到 32,768 字节上限。LD3-base 上同类现象出现在思考档的字符级任务:CUTE 的 del\_word 与 orth 两项分别有 43/100、53/100 的样本顶到 16,384 字节上限;而第 5 节主表的七个榜上,LD3-base 思考档顶上限条数每一项都不多于原版,所以这是局部题型的问题,不是普遍"思考不停"。

\textbf{原因。} 跑飞样本拆开是两种:一半是字节级死循环(\texttt{\textbackslash{}end\{bmatrix\}} 无限重复、连续换行,60 字块重复率 $\geq$ 0.9),一半是 "Let me reconsider… Actually…" 想不完(重复率 $\approx$ 0)。前者是贪心链在字节级更容易锁进周期:每个汉字 3 步、每步 argmax 近乎确定,进入高概率短周期后没有随机性把它带出来。后者是长距离收口从未被监督:结束符都算损失,但没有一个训练窗以算损失的结束符结尾,思考超过 p99 长度后模型从未见过"该结束"——这是暴露偏差,本质是 RL 域的问题。

\textbf{修法。} 不动模型的三条:评测与服务一律采样不贪心;输出上限按字节 ÷ 4 与 BPE 模型对齐(4.3 节);报分时并列"抽不到答案 / 顶上限 / 判不了"计数,因为"抽不到答案按答错记"会把这类样本全算成不会。再加 3.6 节的护栏(预算 + 熔断)。动模型的是 3.4 节的自采 RFT。

\textbf{证据。} 试点线把贪心改成采样后,GPQA 抽不到答案 34 $\rightarrow$ 24 题,分数 54.0 $\rightarrow$ 60.6;再加护栏,思考模式 48.0 $\rightarrow$ 60.6。但三版护栏的答对数全停在 120/198(BPE 原版 170):解码侧只能把"没写完"变成"写完了",写完后对不对取决于模型。

\subsection{按字节采样打断 UTF-8}
\textbf{现象。} 开采样后输出里偶尔冒出不该出现的符号:样本里乘号 \texttt{×} 变成一个几何图形符号、一个上标符号变成一个货币符号(都是与原字符前导字节相同、续字节不同的合法字符)。更坏的情况是直接产生非法字节序列,解码端只能显示替换符。

\textbf{原因。} 一个汉字或符号的三个字节在 BPE 模型里是一次采样,在字节模型里是三次。前导字节之后,温度 0.7 下仍会小概率采到另一个合法续字节(合法但采错),或采到非续字节(非法序列)。前一种是字节采样的固有代价,后一种可以在解码端杜绝。

\textbf{修法。} 解码端加 UTF-8 合法性掩码(3.6 节)。掩码依赖"刚采出的字节"这个状态,所以必须关掉推理框架的重叠调度:否则护栏读到的输出滞后一个 token,前导字节刚出来时护栏仍以为在字符边界,反把合法续字节全禁掉,输出变成"前导、前导、续字节"三连。"合法但采错"的乱码只能靠收紧采样,掩码治不了。

\textbf{证据。} 这个 bug 是在试点线检查点上发现的:同一段中文提示,护栏开 + 重叠调度时输出几乎全是替换符、没有一个完整汉字;护栏全关则偶有非法序列;护栏开 + 关重叠调度后输出全部合法。它一度让我们误以为模型没学会中文编码;所有带护栏的历史分数因此重审。

\subsection{presence / frequency penalty 对 256 词表无效}
\textbf{现象。} 面对复读,通行做法是开 presence / frequency penalty。LD3-base 部署后在真实 agent 输入(1.6 万 token 前缀)上复现复读,加 frequency / presence penalty 无效。

\textbf{原因。} 这两种惩罚按 token id 记账,压低出现过的 id。在 15 万词表里"出现过"有信息量;在 256 个字节里每个常用字节几步之内都出现过,惩罚等于把所有字节一起平抬,压不住 n-gram 级的循环。字节模型的重复抑制必须做在 n-gram 级,推理框架没有现成实现。

\textbf{修法。} 解码侧用 3.6 节的 n-gram 级循环检测,但只喂熔断、平时不改 logits——逐字节禁重复在正常文本上 41\% 的解码步会误触发。训练侧用 3.4 节的 on-policy RFT / DPO。

\textbf{证据。} 见 6.1 的护栏结果与 6.4 的训练结果;penalty 本身没有可报的正面数字。

\subsection{复读及其治理}
\textbf{现象。} LD3-base 自产回答里 46\% 含 8-gram 复读,同题下另一个商用模型 6\%、另一个开源模型 13\%。长上下文(24–32K 前缀)下的中文思考复读是概率事件:原版 Qwen3.8-27B 在同一道中文长任务上 10 次采样也复读 2 次(20\%),采 3 次是 0/3。按更严的口径(循环 / 整句重复)只有 4–6\%,两个数不可直接比。

\textbf{原因。} 三层。第一,字节级把暴露偏差放大约 4 倍,进入短周期后缺少随机性带出(6.1)。第二,原版本来就有一定概率在思考时重复自己;字节化后中文监督字节少(见 3.2 节),中文思考的复读率更高。第三,一部分"磕巴"不是模型的病:客户端流式看到的"你你好""感感文件"是推理框架增量 detokenize 按字符数算偏移、在字节词表上一个汉字三步到齐时重发文本造成的;同一请求非流式输出干净,磕巴率只有 0.2–0.5\%。

\textbf{修法。} 提示无效(LD3-base 不听写进 system 的规范),penalty 无效(6.3),只能训进去:3.4 节的自采(护栏关)$\rightarrow$ RFT $\rightarrow$ 长度归一化 DPO。配对规则决定 DPO 学到什么:chosen 与 rejected 之间除了"病"不能有别的系统差异(长度、思考量、思考语言)。验证复读要采 $\geq$ 10 次报触发率。

\textbf{证据。} M3 链上差样本率(复读 / 磕巴 / 顶上限)11.9\% $\rightarrow$ 7.7\%(直接 DPO)$\rightarrow$ 0.2\%(RFT + 两轮 DPO);代价是精确推理:GSM8K 300 题贪心对话 97.7\% $\rightarrow$ 80.7\%,RFT 先于 DPO 救回到 92.7\%。但 0.2\% 要打折:最后一轮 252 条带思考回答里 211 条思考不到 60 个字符、中位 28 个字符,它大半来自"不思考"。长上下文复读另有结论:SFT 有效(200 条干净长前缀把中文长任务触发率 20\% $\rightarrow$ 0\%;513 条中文 32K 把英文 agent 输入复读 2/10 $\rightarrow$ 1/10),对数不足的 DPO 无效(169 对、218 对各一次)。到 LD3-base 为止复读没有根治,这是第 7 节局限之一。

\section{局限}
\begin{itemize}
\item \textbf{单种子、单次运行}。思考档 0.6 采样的解码随机性没有测;自助法区间只覆盖题目抽样方差。
\item \textbf{bpb 对照与训练数据同分布}(5.2):不能解读为建模能力更强。
\item \textbf{IFEval 非思考档的差距归因不清}:LD3-base 没做过任何指令类后训练,原版做过;本文没有分离这个因素的实验。
\item \textbf{效率是真实代价}:同卡汉字 / 秒约为原版 1/4,长输入 prefill 慢 5–6.5$\times$;草稿头在单卡上没能实测,只能引他处数据。
\item \textbf{字级知识回不全}:读音 / 笔画这类"字 $\rightarrow$ 属性"的查表知识在字节化后下降,与 C-Eval 的损失同向;要多少中文语料才能补回,本文没有回答。
\item \textbf{思考档不稳}:没做后训练的字节模型在部分题型上思考绕不出来(5.4 的 del\_word / orth);治理方法在 3.4 / 6.4 节,但治理后的模型不在本文的对照范围内。
\item \textbf{多语种退化}(5.7):非拉丁 / 西里尔文字的生成退化(保住 11–13/22,原版 21–22/22),复述完好;每种语言只测 3 条,是冒烟测试不是基准,且两列 max\_tokens 数值相同使原版多拿 3–4 倍文本预算。
\item \textbf{视觉塔未重适应};长上下文复读未根治;部分结论(贪心跑飞的量化、惩罚项无效)来自试点线而非 LD3 本身,正文已注明。
\item 评测集与训练集的污染未系统查(C-Eval 题干被用作自建字符题的自然句,但两模型在 E1 中都已评过它,且字符题不考知识)。
\end{itemize}
\section{结论}
把一个 27B 的子词模型原样改成纯字节模型是可行的:词表 248,320 $\rightarrow$ 384,主干不动,4.77 G 字节监督、约 1,430 H200 GPU 时,原版的大部分能力保住了——数学、代码、MMLU-Pro 在噪声内,中文考试与非思考档指令遵循掉 4–6 分。字节化换来的是不靠思考的字符级操作能力(中文尤其明显),付出的是同硬件 1/4–1/5 的字符吞吐与一部分汉字查表知识。字节模型还带来一组自己的失效模式,它们有对应的工程与训练修法。对想在已有模型上尝试字节化的人,我们的建议是三条:对话类监督用目标模型自己的输出;字典(字 $\leftrightarrow$ 字节的显式监督)排在课程最前;评测时按字节 ÷ 4 对齐输出上限并逐题记录顶上限条数。

\appendix\renewcommand{\thetable}{A\arabic{table}}\setcounter{table}{0}\renewcommand{\thefigure}{A\arabic{figure}}\setcounter{figure}{0}
\section{补充表}
\begin{table}[htbp]
\centering
\footnotesize
\caption{字符级任务单项,思考档(每项 100 题,temperature 0.6 / top\_p 0.95,上限 16,384 字节 / 4,096 token;"顶"为顶上限条数)}
\label{tab:e5think}
\begin{adjustbox}{max width=\textwidth}
\setlength{\tabcolsep}{3.5pt}
\begin{tabular}{llrrrrllrrrr}
\toprule
题集 & 项 & orig & \textbf{LD3-base} & $\Delta$ & 顶 orig / LD3 & 题集 & 项 & orig & \textbf{LD3-base} & $\Delta$ & 顶 orig / LD3 \\
\midrule
CUTE & contains\_char & 100 & 100 & 0 & 0 / 0 & 自建 & zh\_count & 93 & 100 & +7 & 0 / 0 \\
CUTE & contains\_word & 100 & 100 & 0 & 0 / 0 & 自建 & zh\_reverse & 59 & 93 & +34 & 0 / 0 \\
CUTE & del\_char & 94 & 98 & +4 & 1 / 2 & 自建 & zh\_nth & 96 & 100 & +4 & 0 / 0 \\
CUTE & del\_word & 64 & 56 & $-$8 & 36 / 43 & 自建 & zh\_index & 95 & 100 & +5 & 1 / 0 \\
CUTE & ins\_char & 85 & 95 & +10 & 2 / 0 & 自建 & zh\_contains & 99 & 100 & +1 & 0 / 0 \\
CUTE & ins\_word & 99 & 100 & +1 & 1 / 0 & 自建 & zh\_del\_char & 94 & 98 & +4 & 0 / 0 \\
CUTE & orth & 67 & 48 & $-$19 & 33 / 53 & 自建 & zh\_ins\_char & 92 & 97 & +5 & 0 / 0 \\
CUTE & sem & 94 & 94 & 0 & 0 / 2 & 自建 & zh\_sub\_char & 88 & 96 & +8 & 1 / 0 \\
CUTE & spell & 99 & 99 & 0 & 1 / 1 & 自建 & zh\_swap\_char & 84 & 98 & +14 & 1 / 0 \\
CUTE & spell\_inverse & 100 & 100 & 0 & 0 / 0 & 自建 & zh\_s2t & 80 & 80 & 0 & 0 / 0 \\
CUTE & sub\_char & 88 & 97 & +9 & 0 / 0 & 自建 & zh\_t2s & 98 & 100 & +2 & 1 / 0 \\
CUTE & sub\_word & 90 & 94 & +4 & 8 / 2 & 自建 & zh\_pinyin & 99 & 88 & $-$11 & 0 / 0 \\
CUTE & swap\_char & 94 & 100 & +6 & 1 / 0 & 自建 & zh\_homophone & 99 & 88 & $-$11 & 1 / 2 \\
CUTE & swap\_word & 77 & 96 & +19 & 23 / 3 & 自建 & zh\_strokes & 69 & 34 & $-$35 & 16 / 4 \\
 &  &  &  &  &  & 自建 & zh\_radical & 72 & 64 & $-$8 & 22 / 14 \\
\bottomrule
\end{tabular}
\end{adjustbox}
\end{table}
\begin{table}[htbp]
\centering
\small
\caption{主表的逐题自助法 95\% 区间(2,000 次重采样;$\Delta$ 为同题配对的 LD3-base $-$ orig;只覆盖题目抽样方差,不覆盖解码随机性)}
\label{tab:boot}
\begin{tabular}{lllll}
\toprule
数据集 & 档 & orig & LD3-base & $\Delta$ [95\% 区间] \\
\midrule
GSM8K & 思考 & 97.0 [96.1, 97.9] & 93.0 [91.7, 94.4] & $-$4.0 [$-$5.3, $-$2.9] \\
MATH-500 & 思考 & 94.0 [91.8, 96.0] & 95.4 [93.4, 97.0] & +1.4 [$-$0.8, +3.6] \\
HumanEval+ & 思考 & 90.2 [85.4, 94.5] & 88.4 [83.5, 93.3] & $-$1.8 [$-$7.3, +3.7] \\
C-Eval & 思考 & 89.9 [88.3, 91.5] & 85.4 [83.4, 87.2] & $-$4.5 [$-$6.3, $-$2.8] \\
MMLU-Pro & 思考 & 80.1 [77.6, 82.5] & 78.8 [76.3, 81.4] & $-$1.3 [$-$3.3, +0.6] \\
LongBench-v2\textdagger{} & 思考 & 27.4 [19.7, 35.9] & 39.3 [30.8, 47.9] & +11.9 [+3.4, +21.4] \\
GSM8K & 非思考 & 96.4 [95.4, 97.3] & 96.0 [94.9, 97.0] & $-$0.4 [$-$1.2, +0.5] \\
MATH-500 & 非思考 & 92.8 [90.4, 95.0] & 93.6 [91.4, 95.6] & +0.8 [$-$1.6, +3.2] \\
HumanEval+ & 非思考 & 92.7 [88.4, 96.3] & 90.2 [85.4, 94.5] & $-$2.5 [$-$6.7, +1.8] \\
C-Eval & 非思考 & 88.1 [86.3, 89.9] & 83.4 [81.4, 85.4] & $-$4.7 [$-$6.5, $-$3.0] \\
MMLU-Pro & 非思考 & 79.4 [76.9, 81.8] & 78.1 [75.6, 80.6] & $-$1.3 [$-$3.3, +0.7] \\
LongBench-v2\textdagger{} & 非思考 & 42.7 [34.2, 52.1] & 53.8 [44.4, 62.4] & +11.1 [+2.6, +19.7] \\
\bottomrule
\end{tabular}
\end{table}

\end{document}